\documentclass[10pt,twocolumn,letterpaper]{article}

\usepackage[pagenumbers]{cvpr} 

\usepackage[utf8]{inputenc} 
\usepackage[T1]{fontenc}    

\usepackage{url}            
\usepackage{booktabs}       
\usepackage{amsfonts}       
\usepackage{nicefrac}       
\usepackage{microtype}      
\usepackage{xcolor}         
\usepackage{subfiles}
\usepackage{bm}
\usepackage{multirow}
\usepackage[accsupp]{axessibility} 
\usepackage{graphicx}
\usepackage{wrapfig}
\usepackage[font=small,labelfont=bf]{caption}
\usepackage{amsmath}
\usepackage{pifont}
\usepackage{colortbl}
\usepackage{tabularx}
\usepackage{arydshln}
\usepackage{makecell}
\usepackage{tikz}

\newcommand*\circled[1]{\tikz[baseline=(char.base)]{
            \node[shape=circle,draw,inner sep=0.6pt] (char) {#1};}}
\usepackage{siunitx}
\newcommand{\ablation}{
\begin{table*}[t]
\caption{Ablation of RPE on ViT-small. Aligned is the standard protocol with raw face images (detector bounding box) aligned by RetinaFace~\cite{deng2020retinaface} and resized to $112\!\times\!112$. Unaligend takes the raw face images and simply resizes it to $112\!\times\!112$. Aligned setting always shows better performances and Unaligned is for simulating alignment failure. Low Quality Aligned dataset may have alignment failures. }
\vspace{1mm}
\centering
\footnotesize
\begin{tabular}{cccccccccc}
\hline
\multicolumn{1}{c}{\multirow{3}{*}{Method}} & \multicolumn{4}{c}{Low Quality Aligned Dataset} & \multicolumn{2}{c}{High Quality Aligned Dataset} & \multicolumn{2}{c}{High Quality Unaligned Dataset} \\ 
& \multicolumn{2}{c}{TinyFace~\cite{tinyface}} & \multicolumn{2}{c}{IJB-S~\cite{ijbs}} & CFPFP~\cite{cfpfp} & IJB-C~\cite{ijbc} & CFPFP~\cite{cfpfp} & IJB-C~\cite{ijbc} \\
& Rank-$1$ & Rank-$5$ & Rank-$1$ & Rank-$5$ & Verification & TAR@0.01\% & Verification & TAR@0.01\% \\ \hline

ViT & $68.24$ & $72.96$ & $59.60$ & $68.31$ & $96.11$ & $92.22$ & $72.81$ & $21.62$ \\ \hline
ViT + iRPE & $69.05$ & $73.10$ & $62.49$ & $70.50$ & $\bm{97.01}$ & $92.72$ & $77.91$ & $34.73$ \\ \hline
ViT+KP-RPE & $\bm{69.88}$ & $\bm{74.25}$ & $\bm{63.44}$ & $\bm{72.04}$ & 96.60 & $\bm{94.20}$ & $\bm{93.56}$ & $\bm{91.85}$ \\ \hline

\end{tabular}

\label{tab:ablation}
\end{table*}

}

\newcommand{\ablationmodules}{
\begin{table*}[t]
\caption{Ablation of KP-RPE with three different formulations of keypoint dependent RPE tables $f(\mathbf{P})$. The sharp increase in Unaligned setting shows the robustness to unseen affine transform manifests with Relative $f(\mathbf{P})$. Multihead $f(\mathbf{P})$ further improves the performance. }
\vspace{1mm}
\centering
\footnotesize
\begin{tabular}{lccccccccc}
\hline
\multicolumn{1}{c}{\multirow{3}{*}{Method}} & \multicolumn{4}{c}{Low Quality Aligned Dataset} & \multicolumn{2}{c}{High Quality Aligned Dataset} & \multicolumn{2}{c}{High Quality Unaligned Dataset} \\ 
& \multicolumn{2}{c}{TinyFace~\cite{tinyface}} & \multicolumn{2}{c}{IJB-S~\cite{ijbs}} & CFPFP~\cite{cfpfp} & IJB-C~\cite{ijbc} & CFPFP~\cite{cfpfp} & IJB-C~\cite{ijbc} \\
& Rank-$1$ & Rank-$5$ & Rank-$1$ & Rank-$5$ & Verification & TAR@0.01\% & Verification & TAR@0.01\% \\ \hline
KP-RPE Absolute $f(\mathbf{P})$                      & $68.11$ & $72.42$ & $9.97$  & $69.13$ & $96.51$ & $90.96$ & $68.09$ & $14.91$ \\ \hline
KP-RPE Relative $f(\mathbf{P})$                   & $69.42$ & $73.71$ & $62.51$ & $70.77$ & $\bm{96.74}$ & $\bm{94.28}$ & $89.70$ & $85.22 $\\ \hline
KP-RPE MultiHead $f(\mathbf{P})$          & $\bm{69.88}$ & $\bm{74.25}$ & $\bm{63.44}$ & $\bm{72.04}$ & $96.60$ & $94.20$ & $\bm{93.56}$ & $\bm{91.85}$ \\ \hline
\end{tabular}

\label{tab:ablationtwo}
\end{table*}

}

\newcommand{\flop}{

\begin{table*}[!t]
    \centering
    
    \caption{Computation resource comparison. GFLOP refers to Giga Floating Operating per Second. We measure it as~\cite{rasley2020deepspeed}. Throughput refers to the number of images processed per second during the train/eval iteration. }
    \footnotesize
    \begin{tabular}{lccrrcc}
\hline
     & GFLOP & $\Delta$ in GFLOP &  \makecell{Eval\\Throughput} & \makecell{Train\\Throughput} & \makecell{ \%$\Delta$ in  Train \\Throughput} & \# Param \\ \hline
IResNet50 & $12.62$ & - & $1432.72$ img/s & $337.93$ img/s & - & $43.59$M \\ \hline
ViT Small  & $17.42$ & \circled{1} &  $1303.15$ imgs/s & $333.17$ img/s & \circled{1} & $95.95$M \\ 
ViT Small + iRPE  & $18.13$ & \circled{1}+$0.71$ & $832.12$ imgs/s & $186.55$ img/s & \circled{1}-$44.01$\% & $96.07$M \\ 
ViT Small + \textbf{KP-RPE}  & $17.44$ & \circled{1}+$0.02$ & $1145.90$ imgs/s & $302.70$ img/s & \circled{1}-$9.15$\% & $96.00$M \\ 
ViT Small + \textbf{KP-RPE (+ Ldmk) } & $17.58$ & \circled{1}+$0.16$ & $1085.22$ imgs/s & $302.70$ img/s & \circled{1}-$9.15$\% & $96.49$M \\ \hline\hline
IResNet101  & $24.19$ & - &  $773.12$  imgs/s & $189.74$ img/s & - & $65.15$M \\ \hline
ViT  Base  & $24.83$ & \circled{2} & $644.10$ imgs/s  & $162.94$ img/s & \circled{2} & $114.87$M \\ 
ViT  Base + iRPE  & $26.25$ & \circled{2}+$1.42$ & $337.32$ imgs/s   & $79.40$ img/s & \circled{2}-$51.27$\% & $114.98$M \\
ViT  Base + \textbf{KP-RPE}  & $24.90$ & \circled{2}+$0.07$ & $502.57$ imgs/s  & $136.15$ img/s & \circled{2}-$16.44$\% & $115.08$M \\
ViT  Base + \textbf{KP-RPE (+ Ldmk)}  & $25.04$ & \circled{2}+$0.21$ &  $489.37$ imgs/s & $136.15$ img/s & \circled{2}-$16.44$\% & $115.56$M \\ \hline
\end{tabular}

    \label{tab:flop}
\end{table*}
}

\newcommand{\sotatable}{

\begin{table*}[!t]
\centering

\caption{SoTA comparison on low-quality and high-quality datasets. ViT models are ViT-Base sized.} 
\vspace{-1mm}
\footnotesize
\label{SOTA_table}
\scalebox{1.0}{
\begin{tabular}{lccccccccc}
\hline
\multicolumn{1}{c}{\multirow{3}{*}{Method}} & \multicolumn{1}{c}{\multirow{3}{*}{Backbone}} & \multicolumn{1}{c}{\multirow{3}{*}{Train Data}}  &\multicolumn{4}{c}{Low Quality Dataset}   &\multicolumn{3}{c}{High Quality Dataset} \\ 
   &  &  & \multicolumn{2}{c}{TinyFace~\cite{tinyface}}   & \multicolumn{2}{c}{IJB-S~\cite{ijbs}}    &  AgeDB~\cite{agedb}  &  CFPFP~\cite{cfpfp} &  IJB-C~\cite{ijbc}  \\
           &   &   &   \multicolumn{1}{c}{Rank-$1$} & \multicolumn{1}{c}{Rank-$5$} & \multicolumn{1}{c}{Rank-$1$} & \multicolumn{1}{c}{Rank-$5$}  & \multicolumn{2}{c}{\scalebox{0.875}{Verification Accuracy}} & \scalebox{0.875}{TAR@FAR=0.01\%} \\ \hline
PFE~\cite{shi2019probabilistic}\scalebox{0.875}{\textcolor{white}{aaa}}  &  CNN64  & MS1MV2~\cite{deng2019arcface}         & -  & -         & $50.16$         & $58.33$    & -  & -  & -        \\
ArcFace~\cite{deng2019arcface}    & ResNet101 & MS1MV2~\cite{deng2019arcface}         & - & -       & $57.35$         & $64.42$   & $98.28$ & $98.27$ &   $96.03$      \\
URL~\cite{shi2020towards}    &    ResNet101    & MS1MV2~\cite{deng2019arcface}          & ${63.89}$ &  ${68.67}$      & $59.79$         & $65.78$        & - & $98.64$ &   $96.60$     \\
\scalebox{0.86}{CurricularFace~\cite{huang2020curricularface}} & ResNet101  & MS1MV2~\cite{deng2019arcface}       & $63.68$ & $67.65$        & ${62.43}$         & ${68.68}$    & $\mathbf{98.32}$  & $98.37$ &     $96.10$    \\ 
AdaFace~\cite{deng2019arcface}    &   ResNet101    & MS1MV2~\cite{deng2019arcface}          &  ${68.21}$ & ${71.54}$     & ${65.26}$         & ${70.53}$     & $98.05$ & $98.49$ &       $96.89$   \\
AdaFace~\cite{deng2019arcface}    &   ResNet101    & MS1MV3~\cite{deng2019lightweight}        &  $67.81$  & $70.98$        & $67.12$         & $72.67$     & $98.17$ & $99.03$ &  $97.09$     \\\hdashline
AdaFace~\cite{kim2022adaface}   &  ViT    &  MS1MV3~\cite{deng2019lightweight}  & $72.05$	 & $74.84$  & $65.95$  & $71.64$  & $97.87$  &  $99.06$ & $97.10$\\    
AdaFace~\cite{kim2022adaface}   &  \textbf{ViT+KP-RPE}    &  MS1MV3~\cite{deng2019lightweight}  & $\mathbf{73.50}$   &   $\mathbf{76.39}$   & $\mathbf{67.62}$ &  $\mathbf{73.25}$   & $97.98$ & $\mathbf{99.11}$ &    $\mathbf{97.16}$
\\\hline 
ArcFace~\cite{deng2019arcface}   &   ResNet101   & \scalebox{0.86}{WebFace4M~\cite{zhu2021webface260m}}   &  $71.11$  &  $74.38$   & $69.26$ & $74.31$   & $\mathbf{97.93}$ & $99.06$ & $96.63$ \\
AdaFace~\cite{kim2022adaface}   &    ResNet101   &  \scalebox{0.86}{WebFace4M~\cite{zhu2021webface260m}}   &  ${72.02}$  &  ${74.52}$   &  ${70.42}$ & ${75.29}$   & $97.90$  & $\mathbf{99.17}$ &  $\mathbf{97.39}$  \\\hdashline
AdaFace~\cite{kim2022adaface}   &  ViT    &  \scalebox{0.86}{WebFace4M~\cite{zhu2021webface260m}}    & $74.81$ &  $77.58$ & $71.90$  & $77.09$  & $97.48$  &  $98.94$ & $97.14$ \\    
AdaFace~\cite{kim2022adaface}   &  ViT+iRPE    &  \scalebox{0.86}{WebFace4M~\cite{zhu2021webface260m}}   &  $74.92$  &  $77.98$   & $71.93$  & $77.14$  & $97.15$ & $99.01$ &  $97.01$ \\
AdaFace~\cite{kim2022adaface}   &  \textbf{ViT+KP-RPE}    &  \scalebox{0.86}{WebFace4M~\cite{zhu2021webface260m}}   &  $\mathbf{75.80}$  &  $\mathbf{78.49}$   & $\mathbf{72.78}$  & $\mathbf{78.20}$  & $97.67$ & $99.01$ &  $97.13$ 
\\\hline
AdaFace~\cite{kim2022adaface}   &    ResNet101   &  \scalebox{0.86}{WebFace12M~\cite{zhu2021webface260m}}   &  $72.42$  &  $74.81$   &  $71.46$ & $77.04$   & $98.00$ & $99.24$ &  $97.66$  \\\hdashline
AdaFace~\cite{kim2022adaface}   &  \textbf{ViT+KP-RPE}    &  \scalebox{0.86}{WebFace12M~\cite{zhu2021webface260m}}   & $\mathbf{76.18}$ & $\mathbf{78.97}$  & $\mathbf{72.94}$  & $\mathbf{77.46}$  &  $\mathbf{98.07}$ & $\mathbf{99.30}$  & $\mathbf{97.82}$\\\hline
\end{tabular}
}
\small

\end{table*}
}

\newcommand{\figurezero}{
\begin{figure}[t]
    \centering
    \includegraphics[width=\linewidth]{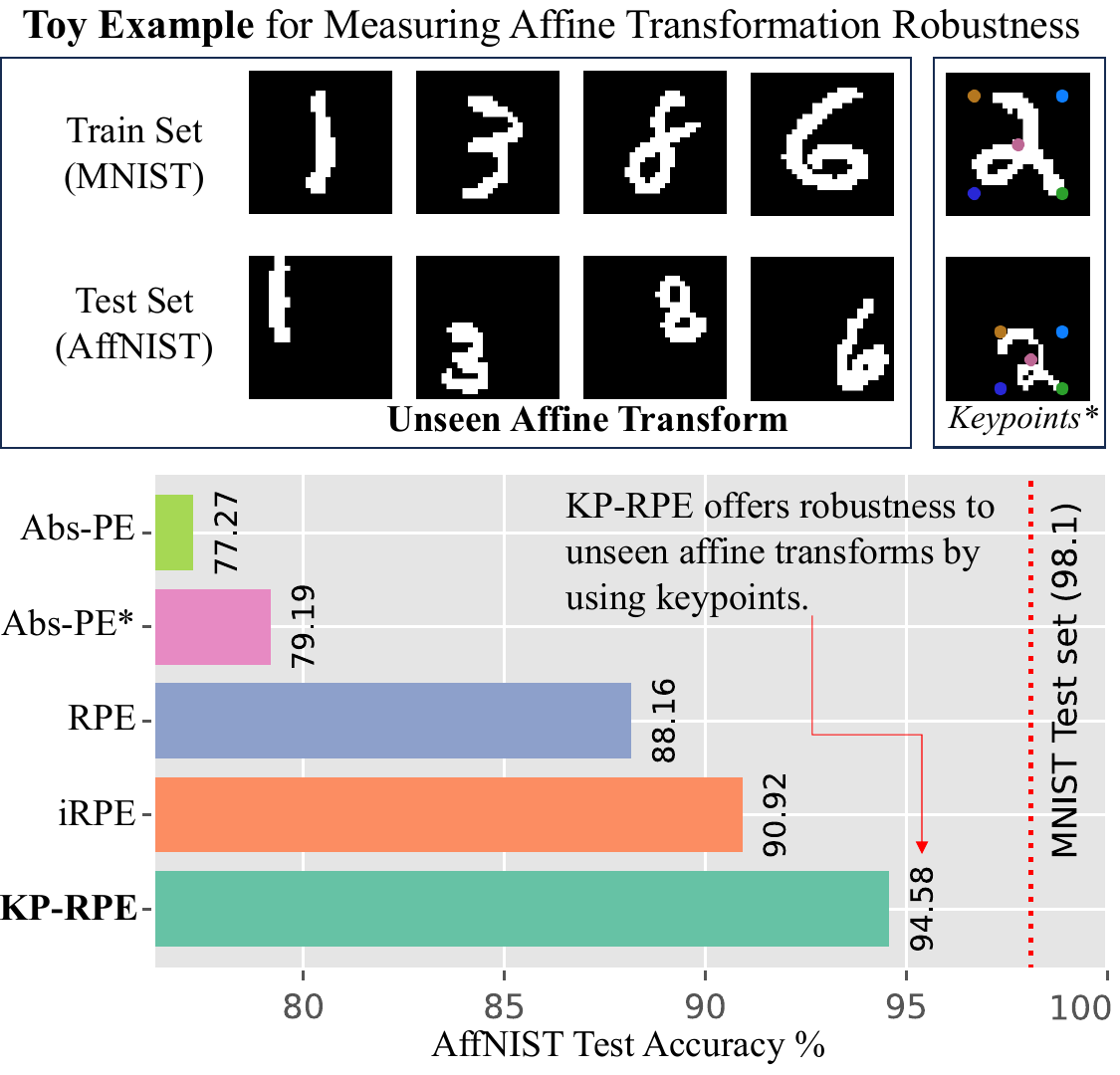}
    \caption{Toy Example illustrating how different Position Embeddings impact the ViT's robustness to unseen affine transforms. Abs-PE refers to the learnable Absolute Position Embedding. RPE and iRPE refers to Relative Position Embedding adopted to ViT~\cite{huang-etal-2020-improve,wu2021rethinking}. Keypoints in MNIST is arbitrarily defined to be the four corners of a box that covers a digit. Abs-PE* is drawing the keypoints onto the input image. KP-RPE uses the keypoints to adjust the RPE. }
    \label{fig:zero}
\end{figure}
}

\newcommand{\figureone}{
\begin{figure*}[h]
    \centering
    \includegraphics[width=\linewidth]{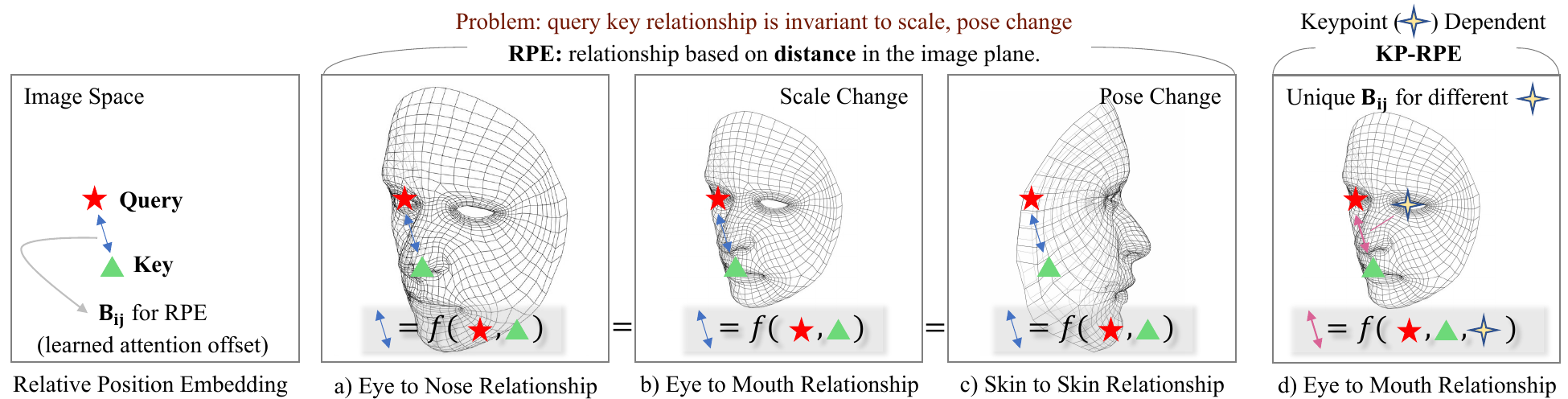}
    \caption{Illustration of RPE~\cite{shaw2018self} and proposed KP-RPE. The blue arrow represents the learned attention offset $\mathbf{B}_{ij}$ between a query $i$ and key $j$ of attention in RPE. The query-key relationship at the same $i$ and $j$ should represent different relationships as the scale or pose change. But $\mathbf{B}_{ij}$ does not change in RPE. KP-RPE addresses this issue by incorporating the \textit{distance to the keypoints} when calculating the learned attention offset in RPE. }
    \label{fig:figure1}
\end{figure*}
}

\newcommand{\figuretwo}{
\begin{figure}
    \centering
    \includegraphics[width=\linewidth]{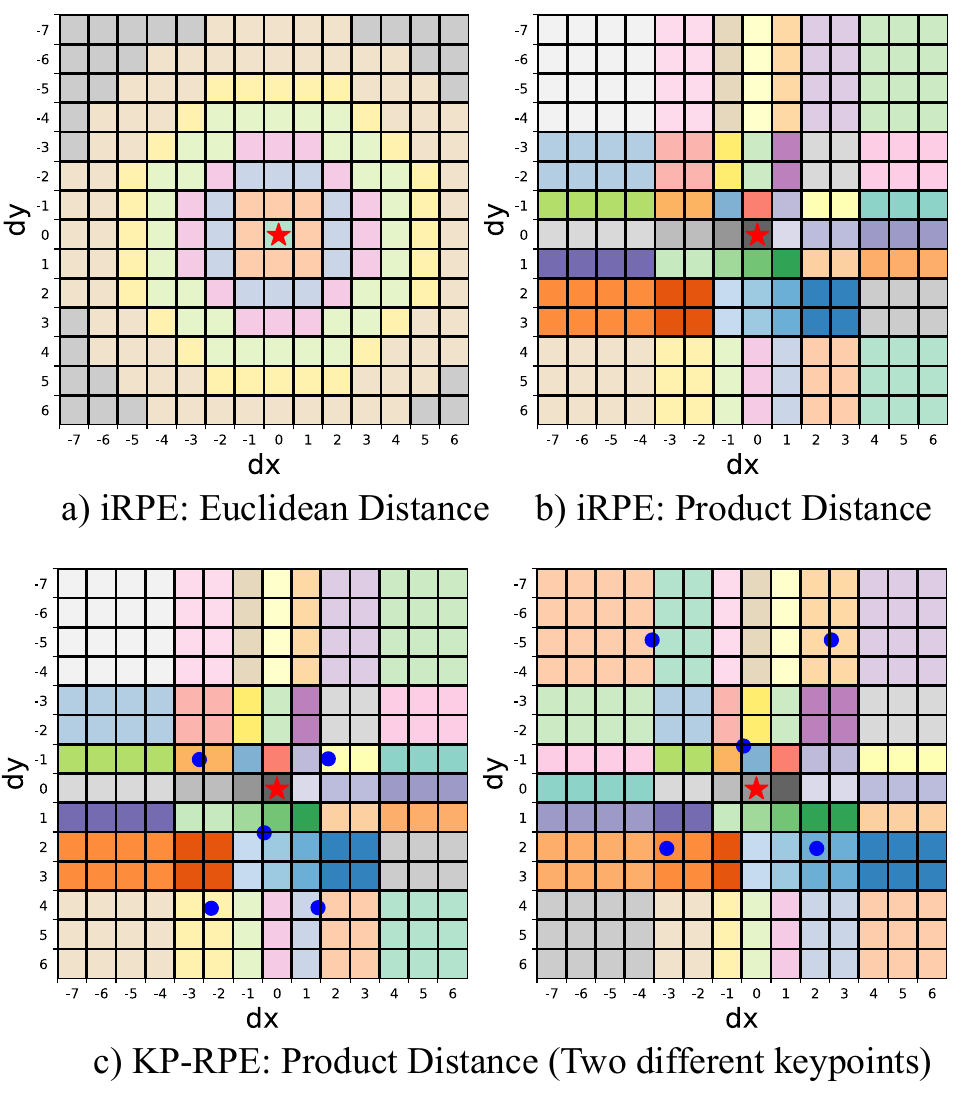}
    \caption{Depiction of key-query combinations in an image, given a query location $i=(7,7)$ (\textcolor{red}{$\star$}). Distinct colors represent varying attention offset values in RPE based on the distance between $i$ and $j$. We are showing $\mathbf{B}_{i=(7,7),j}$ for all $j\in{(14\times14})$. a) The distance function is a quantized Euclidean distance. b) Product distance proposed in iRPE accounts for direction. c) We adopt b) and allow $\mathbf{B}_{i,j}$ to vary based on keypoint locations (\textcolor{blue}{$\bullet$}). }
    \label{fig:figure2}
\end{figure}
}

\newcommand{\figurethree}{
\begin{figure*}
\vspace{-2mm}
    \centering
    \includegraphics[width=\linewidth]{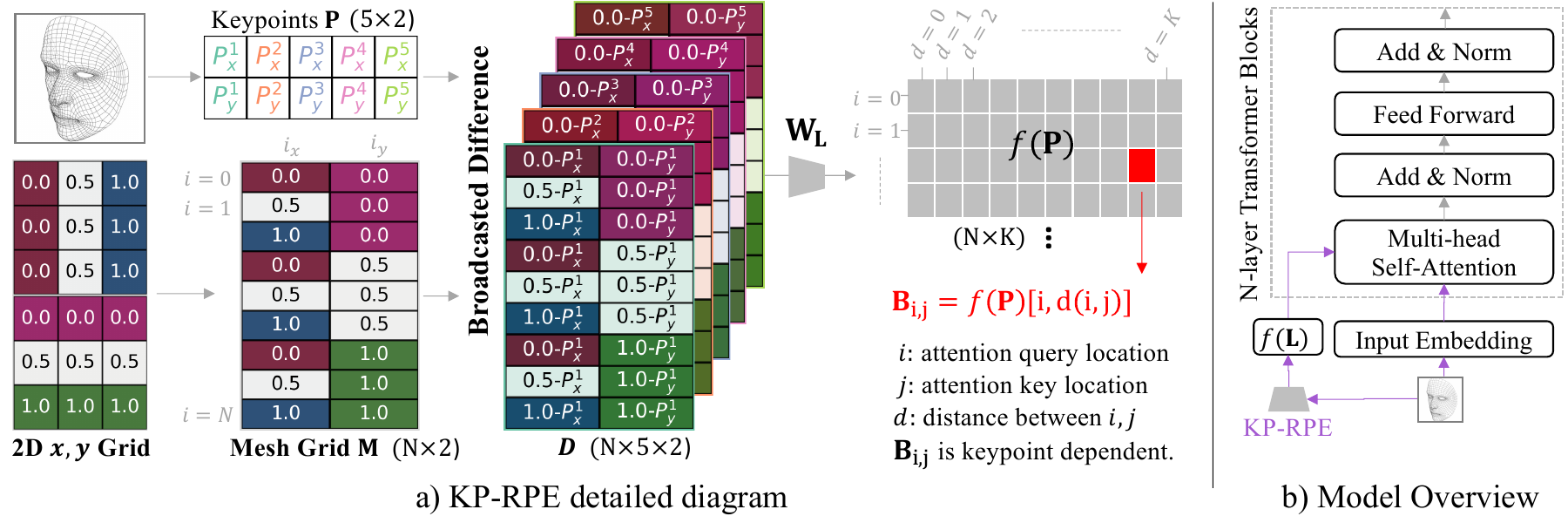}
    \caption{a) Illustration of KP-RPE. First a mesh grid $\mathbf{M}$ and an image-specific keypoints $\mathbf{P}$ are generated. Then the broadcasted difference $\mathbf{D}$ is calculated, and we linearly map $\mathbf{D}$ to $f(\mathbf{P})$. Finally for a given $i,j$, we can find the $\mathbf{B}_{ij}=f(\mathbf{P})[i, d(i,j)])$, which is used to adjust the attention map in self-attention. b) Backbone contains multiple transformer blocks followed by an MLP for classification. KP-RPE is used where multi-head attention modules exist.
    KP-RPE is efficient as $f(\mathbf{P})$ is computed once.}
    \label{fig:figure3}
    \vspace{-2mm}
\end{figure*}
}

\newcommand{\interoplation}{
\begin{figure}
\centering
  \includegraphics[width=0.4\textwidth]{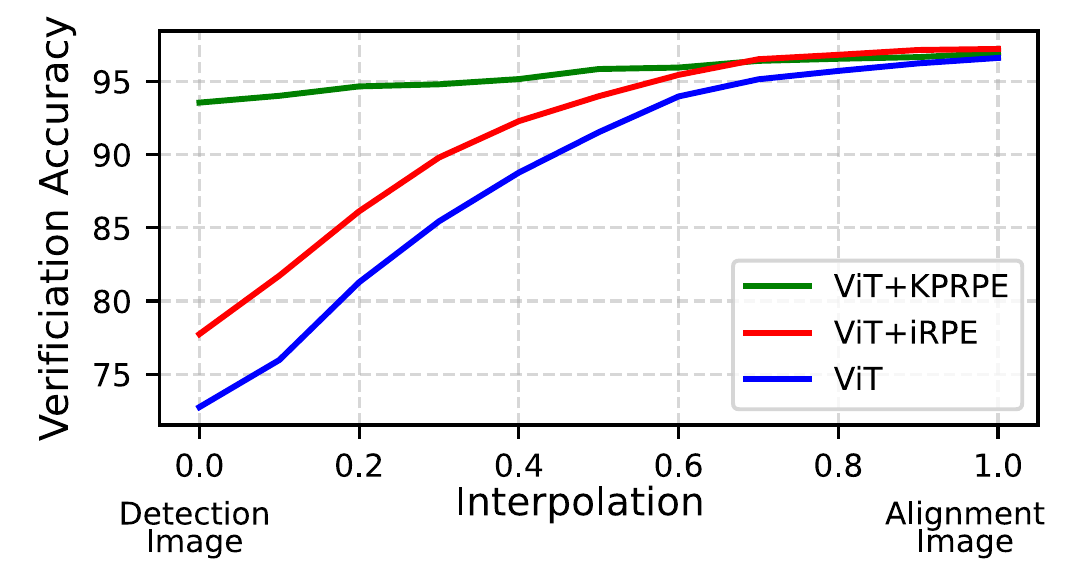}
  \caption{Plot of Verification Accuracy in CFPFP~\cite{cfpfp}. On the X-axis, we interpolate the affine transformation from raw data (Detection Image) to canonical alignment (Alignment Image). Note KP-RPE is robust to affine transformations, while all models have been trained on the aligned image dataset. 
  }
  \label{fig:interpolation}
  \end{figure}
}

\newcommand{\Paragraph}[1]{\vspace{1mm}\noindent\textbf{#1.}\hspace{0.5mm}}

\definecolor{mygreen}{rgb}{0.0, 0.8, 0.6}
\newcommand{\makegreen}[1]{\textcolor{mygreen}{#1}}

\definecolor{cvprblue}{rgb}{0.21,0.49,0.74}
\usepackage[pagebackref,breaklinks,colorlinks,citecolor=cvprblue]{hyperref}

\def\paperID{3057} 
\def\confName{CVPR}
\def\confYear{2024}

\title{KeyPoint Relative Position Encoding for Face Recognition}

\author{
Minchul Kim\quad\quad Yiyang Su\quad\quad Feng Liu\quad\quad Anil Jain\quad\quad Xiaoming Liu \\
{\tt\small \{kimminc2; suyiyan1; liufeng6; jain; liuxm\}@msu.edu} \\
}

\begin{document}

\maketitle

\begin{abstract}
In this paper, we address the challenge of making ViT models more robust to unseen affine transformations. Such robustness becomes useful in various recognition tasks such as face recognition when image alignment failures occur. We propose a novel method called KP-RPE, which leverages key points (e.g.~facial landmarks) to make ViT more resilient to scale, translation, and pose variations. We begin with the observation that Relative Position Encoding (RPE) is a good way to bring affine transform generalization to ViTs. RPE, however, can only inject the model with prior knowledge that nearby pixels are more important than far pixels. Keypoint RPE (KP-RPE) is an extension of this principle, where the significance of pixels is not solely dictated by their proximity but also by their relative positions to specific keypoints within the image. By anchoring the significance of pixels around keypoints, the model can more effectively retain spatial relationships, even when those relationships are disrupted by affine transformations. We show the merit of KP-RPE in face and gait recognition. The experimental results demonstrate the effectiveness in improving face recognition performance from low-quality images, particularly where alignment is prone to failure.
Code and pre-trained models are \href{https://github.com/mk-minchul/kprpe}{available}.
\end{abstract}

\section{Introduction}
Geometric alignment has shown to be highly effective for certain computer vision problems, such as face, body and gait recognition~\cite{deng2019arcface,liu2017sphereface,wang2018cosface,kim2022adaface,meng2021magface,deng2021variational,kim2020broadface,li2021spherical,zhao2023grouped,agedb,cfpfp,lfw,cplfw,ijba,ijbb,ijbc,zheng2022gait,zhang2020learning,liuiccv2023}. Alignment is the process of transforming input images, to a consistent and standardized form, often by 
scaling, rotating, and translating.
This standardization helps recognition models learn the underlying patterns and features more effectively. As a result, many state-of-the-art (SoTA) face recognition models~\cite{deng2019arcface,kim2022adaface,meng2021magface,wang2018cosface} rely on well-aligned datasets~\cite{zhu2021webface260m,msceleb,deng2019arcface,deng2020retinaface} to achieve high accuracy.

\figurezero


Fig.~\ref{fig:zero} shows a toy example with a training dataset MNIST~\cite{deng2012mnist} and test set AffNIST~\cite{affnist} which is in unseen affine transformation of MNIST. Using a shallow ViT~\cite{vit} model, one can easily achieve $98.1\%$ accuracy in the MNIST test set. However, in AffNIST, ViT with the original Absolute Position Embedding obtains only $77.27\%$ accuracy. Such a sharp decrease in performance with unseen affine transform causes problems in applications that rely on accurate input alignments.   

\figureone

In face recognition, alignment can be imperfect, especially in low-quality images where accurate landmark detection is difficult~\cite{deng2020retinaface,farsight-a-physics-driven-whole-body-biometric-system-at-large-distance-and-altitude}. Thus, images with low resolution or taken in poor lighting may result in misalignment during testing. 
Given the interplay between alignment and recognition, it becomes crucial to proactively handle potential alignment failures, which often result from, \textit{e.g.}, low-quality images. 
In other words, there is a need for a recognition model that is robust to scale, rotation, and translation variations.



We revisit the Relative Position Encoding (RPE) concept used in ViT~\cite{vit} and find that RPE can be useful for introducing affine transform robustness. RPE~\cite{shaw2018self} enables the model to capture the relative spatial relationships among regions of an image, learning the positional dependencies without relying on absolute coordinates.  As shown in Fig.~\ref{fig:zero}, adding RPE to ViT increases the performance in AffNIST.  With RPE~\cite{shaw2018self}, queries and keys of self-attention~\cite{vaswani2017attention} at closer distances can be assigned different attention weights compared to those at a greater distance.
While RPE allows the model to exploit relative positions, it has a limitation: even if an image changes in terms of scaling, shifting, or orientation, the significance of the key-query position in RPE stays the same. This static behavior is illustrated in Figs.~\ref{fig:figure1} a)-c). Notably, the key-query relationship is the same regardless of the corresponding pixels' semantic meaning changes.

We hypothesize that an RPE which dynamically adapts based on image keypoints, such as facial landmarks, could improve the model's comprehension of spatial relationships in the image. By leveraging the spatial relationships with respect to these keypoints, the model can adapt to variations in scale, rotation, and translation, resulting in a more robust recognition system capable of handling both aligned and misaligned datasets. Fig.~\ref{fig:figure1} d) highlights a keypoint-dependent query-key relationship.

To this end, we introduce KeyPoint RPE (KP-RPE), a method that dynamically adapts the spatial relationship in ViT based on the keypoints present in the image. Our experiments demonstrate that incorporating KP-RPE into ViT significantly enhances the model's robustness to misaligned test datasets while maintaining or even improving performance on aligned test datasets. We show the usage of KP-RPE in face recognition and gait recognition as the inputs share the same topology (face or body) that allows the keypoints to be defined.
Finally, KP-RPE is an order of magnitude faster than iRPE~\cite{wu2021rethinking}, a widely used RPE that depends on the image content. 

In summary, the contributions of this paper include:
\begin{itemize}
    \item The insight that RPE (or its variants) can improve the robustness of ViT to unseen affine transformations. 
    \item  The development of Keypoint RPE (KP-RPE), a novel method that dynamically adapts the spatial relationship in Vision Transformers (ViT) based on the keypoints in the image, significantly enhancing the model's robustness to misaligned test datasets while maintaining or improving performance on aligned test datasets.
    \item Comprehensive experimental validation demonstrating the effectiveness of our proposed KP-RPE, showcasing its potential for advancing the field of recognition by bringing model's robustness to geometric transformation. We improve the recognition performance across unconstrained face datasets such as TinyFace~\cite{tinyface} and IJB-S~\cite{ijbs} and even non-face datasets such as Gait3D~\cite{zheng2022gait,fan2022opengait}.
    
\end{itemize}

\section{Related Works}

\paragraph{Relative Position Encoding in ViT}
Relative Position Encoding (RPE) is first introduced by Shaw \textit{et al}.~\cite{shaw2018self} as a technique for encoding spatial relationships between different elements in a sequence. By adding relative position encodings into the queries and keys, the model can effectively learn positional dependencies without relying on absolute coordinates. Subsequent works, such as those by Dai \textit{et~al}.~\cite{dai2019transformer} and Huang \textit{et~al}.~\cite{huang-etal-2020-improve}, refine and expand upon the concept of RPE, demonstrating its effectiveness in natural language processing (NLP) tasks.

The adoption of RPE in Vision Transformers~\cite{vit} has been explored by several researchers. For instance, Ramachandran \textit{et al}.~\cite{ramachandran2019stand} propose a 2D RPE method that computes the $x$, $y$ distance in an image plane separately to include directional information. A notable RPE method in ViT is iRPE~\cite{wu2021rethinking}, which considers directional relative distance modeling as well as the interactions between queries and relative position encodings in a self-attention mechanism.

Despite the success of these RPE methods in various vision tasks, they do not specifically address the challenges associated with scale, rotation, and translation variations in face recognition applications. This shortcoming highlights the need for RPE methods that can better handle these variations, which are common in real-world low-quality face recognition scenarios. To address this, we propose KP-RPE, which incorporates keypoint information during the network's feature extraction, significantly enhancing the model's ability to generalize across affine transformations.

\paragraph{Face Recognition with Low-Quality Images}
Recent FR models~\cite{wang2018cosface, deng2019arcface, liu2017sphereface, huang2020curricularface,meng2021magface} have achieved high performance on datasets with discernable facial attributes, such as LFW~\cite{lfw}, CFP-FP~\cite{cfpfp}, CPLFW~\cite{cplfw}, AgeDB~\cite{agedb}, CALFW~\cite{calfw}, IJB-B~\cite{ijbb}, and IJB-C~\cite{ijbc}. However, unconstrained scenarios like surveillance or low-quality videos present additional challenges. Datasets in this setting, such as TinyFace~\cite{tinyface} and IJB-S~\cite{ijbs}, contain low-quality images, 
where detecting facial landmarks and achieving proper alignment becomes increasingly difficult. 
This adversely affects existing FR models that rely on well-aligned inputs.

 Several studies~\cite{ranjan2018crystal,zheng2020automatic,zheng2019uncertainty,liu2022controllable,kim2022adaface,shin2022teaching,grosz2023latent,grm2023meet,kuo2022towards,kim2022caface,huang2020improving,yin2020fan} tackle recognition with low-quality imagery. Particularly, AdaFace~\cite{kim2022adaface} introduces an image quality adaptive loss function that reduces the influence of low-quality or unidentifiable samples. 
 A-SKD~\cite{shin2022teaching} employs teacher-student distillation to focus on similar areas regardless of image resolution. But, these models, which are trained on aligned training sets, do not tackle the challenges associated with misaligned inputs in real-world situations. In contrast, KP-RPE adjusts spatial relationships within ViT based on image keypoints, allowing the model to better generalize even when alignment is unsuccessful in low-quality imagery.




\paragraph{Keypoints and Spatial Reasoning}
Keypoint detection, often associated with landmarks, has been fundamental in various vision tasks such as human pose estimation~\cite{cao2017realtime, newell2016stacked}, face landmark detection~\cite{zhang2016joint, bulat2017far,tai2019towards,luvli-face-alignment-estimating-landmarks-location-uncertainty-and-visibility-likelihood}, and object localization~\cite{papandreou2017towards}. These keypoints serve as representative points that capture the essential structure or layout of an object, facilitating tasks like alignment, recognition, and even animation.

Face landmark detection is commonly carried out alongside face detection. MTCNN~\cite{zhang2016joint} is a widely-used method for combined face detection and facial landmark localization, utilizing cascaded CNNs (P-Net, R-Net, and O-Net) that collaborate to detect faces and landmarks in an image. RetinaFace~\cite{deng2020retinaface}, on the other hand, is a single-stage detector~\cite{liu2016ssd,fpn} based landmark localization algorithm, demonstrating strong performance when trained on the annotated WiderFace~\cite{yang2016wider} dataset. TinaFace~\cite{zhu2020tinaface} further enhances detection capabilities by incorporating SoTA generic object detection algorithms. MTCNN and RetinaFace are often used for aligning face datasets.

Recent advances in keypoint detection techniques, particularly using deep neural networks, have led to using keypoints to improve the performance of recognition tasks~\cite{yan2018spatial, su2020human}. For instance, ~\cite{hachiuma2023unified} proposes a keypoint-based pooling mechanism and shows promising results in skeleton-based action recognition and spatio-temporal action localization tasks. Albeit its benefit, many models including ViTs do not have pooling mechanisms. KP-RPE is the first attempt at incorporating keypoints into the RPE which can be easily inserted into ViT models.



\section{Proposed Method}

\subsection{Background}
\paragraph{Self-Attention}
Self-attention is a crucial component of transformers~\cite{vaswani2017attention}, which is a popular choice for a wide range of NLP tasks. ViT~\cite{vit} applies the same self-attention mechanism to images, treating images as sequences of non-overlapping patches. The self-attention mechanism in Transformers calculates attention weights based on the compatibility between a query and a set of keys. Given a set of input vectors, the Transformer computes query ($\mathbf{Q}$), key ($\mathbf{K}$), and value ($\mathbf{V}$) matrices through linear transformations:
\begin{align}
\mathbf{Q}_i = \bm{x_i} \mathbf{W}_Q, \quad
\mathbf{K}_j = \bm{x_j} \mathbf{W}_K, \quad
\mathbf{V}_j = \bm{x_j} \mathbf{W}_V,
\end{align}
where $\bm{x_i}$ is the $i$-th input vector, and $\mathbf{W}_Q$, $\mathbf{W}_K$, and $\mathbf{W}_V$ are learnable weight matrices.

The self-attention mechanism computes attention weights as the dot product between the query and key vectors, followed by a softmax normalization:
\begin{align}
e_{ij} = \frac{\mathbf{Q}_i \mathbf{K}_j^T}{\sqrt{d_k}}, \quad
a_{ij} = \frac{\exp(e_{ij})}{\sum_{j=1}^N \exp(e_{ij})},
\end{align}
where $d_k$ is the dimension of the key vectors. Finally, the output matrix $\mathbf{Y}$ is computed as the product of the attention weight matrix and the value matrix: $\mathbf{Y}_i = \sum_{j=1}^{N} a_{ij} \mathbf{V}_j$.

\paragraph{Absolute Position Encoding}

Transformers are inherently order invariant, as their self-attention mechanism does not consider input token positions. To address this, absolute position encoding is introduced~\cite{vaswani2017attention,gehring2017convolutional}, which adds fixed, learnable positional embeddings to input tokens:
\begin{equation}
\bm{x_i}^{'} = \bm{x_i} + \makegreen{\text{PE}(i)},
\end{equation}
where $\bm{x_i}^{'}$ is the updated input token with positional information, $\bm{x_i}$ is the original input token, and $\text{PE}(i)$ is the positional encoding for the $i$-th position. These embeddings, generated using sinusoidal functions or learned directly, enable the model to capture the absolute positions of elements.

\paragraph{Relative Position Encoding (RPE)}

RPE, introduced by Shaw \textit{et al}.~\cite{shaw2018self} and refined by Dai \textit{et al}.~\cite{dai2019transformer} and Huang \textit{et al}.~\cite{huang-etal-2020-improve}, encodes relative position information, essential for tasks focusing on input element relationships. Unlike absolute position encoding, RPE considers query-key interactions based on sequence-relative distances. The modified self-attention calculation for RPE is:
\begin{equation}
e_{ij}^\prime = \frac{(\mathbf{Q}_i + \makegreen{\mathbf{R}^Q_{ij}}) (\mathbf{K}_j + \makegreen{\mathbf{R}^K_{ij}})^T}{\sqrt{d_k}}, \mathbf{Y}_i = \sum_{j=1}^{n} a_{ij} (\mathbf{V}_j + \makegreen{\mathbf{R}^V_{ij}}).
\end{equation}
Here, $\mathbf{R}^Q_{ij}$, $\mathbf{R}^K_{ij}$, and $\mathbf{R}^V_{ij}$ are relative position encoding between the $i$-th query and $j$-th key with shape $\mathbb{R}^{d_z}$. Each $\mathbf{R}$ is a learnable matrix of $\mathbb{R}^{K\times d_z}$, where $\mathbf{R}_{i,j}$ corresponds to the relative position encoding for distance $d(i, j)=k$ and $K$ is the maximum possible value of $d(i, j)$. To obtain relative position encoding, we index the $\mathbf{R}$ matrix using the computed distance $\mathbf{R}[d(i, j)]$. Common choices for $d$ are quantized Euclidean distance, separate $x,y$ cross distance~\cite{ramachandran2019stand}. \cite{wu2021rethinking} uses a quantized $x,y$ product distance, which encodes direction information. Note, query location $i$ is a 2D point ($i_x, i_y$). Fig.~\ref{fig:figure2} a) and b) illustrate the distance between $i$ and all possible $j$ with different distance functions.  For KP-RPE, we modify~\cite{wu2021rethinking} and allow the RPE to be keypoint dependent.

\figuretwo

\subsection{Keypoint Relative Position Encoding}
\label{sec:kprpe}
Building upon the general formulation of~\cite{wu2021rethinking}, we begin with the following RPE formulation:
\begin{equation}
e_{ij}^\prime = \frac{\mathbf{Q}i \mathbf{K}j^T + \makegreen{\mathbf{B}_{ij}} }{\sqrt{d_k}}.
\end{equation}
Here, $\mathbf{B}_{ij}$ is a scalar that adjusts the attention matrix based on the query and key indices $i,j$. Assuming a set of keypoints $\mathbf{P}\in \mathbb{R}^{N_L\times 2}$ is available for each $\bm{x}$, our goal is to make $\mathbf{B}_{ij}$ dependent on $\mathbf{P}$. For face recognition, $\mathbf{P}$ is the five facial landmarks (two eyes, nose, mouth tips). For gait recognition, $\mathbf{P}$ is 17 points from the joint locations of skeleton predictions. For the MNIST toy example, $\mathbf{P}$ is five keypoints from the four corners and the center of the minimum cover box of a foreground image. As such $\mathbf{P}$ can be defined for objects with shared topology.  

\figurethree

The novelty of KP-RPE lies in the design of $\mathbf{B}_{ij}$. Since
\begin{equation}
    \mathbf{B}_{ij} = \mathbf{W}[d(i,j)] \in \mathbb{R}^{1},
\end{equation}
comprises of a learnable table $\mathbf{W}$ and a distance function $d(i,j)$, we can make $\mathbf{W}$ or $d(i,j)$ depend on the keypoints. At a first glance, $d(i,j,\mathbf{P})$, conditioning the distance on $\mathbf{P}$ seems plausible. 
However, we find that it leads to inefficiencies, as distance caching, which is precomputing $d(i,j)$ for a given input size, is only feasible when $d(i,j)$ is independent of the input. Therefore, we propose an alternative where the bias matrix itself,  $\mathbf{W}$, is a function of $\mathbf{P}$:
\begin{equation}
\mathbf{B}_{ij} = f(\mathbf{P})[d(i,j)].
\end{equation}
We propose three variants of $f(\mathbf{P})$ building up from the simplest solution. 

\Paragraph{Absolute $f(\mathbf{P})$}
Let $\mathbf{P}\in\mathbb{R}^{N_L\times 2}$ be the normalized keypoints between 0 and 1. First, the simplest way to model the indexing table is to linearly map $\mathbf{P}$ to the desired shape. $f(\mathbf{P}) = \mathbf{P}^{\prime} \mathbf{W}_L$ where $\mathbf{P}^{\prime} \in \mathbb{R}^{1\!\times\!(2N_L)}$ is reshaped keypoints $\mathbf{P}$ and $\mathbf{W}_L\in \mathbb{R}^{(2N_L)\times K}$ is a learnable matrix. $K$ is the maximum distance value in $d(i,j)$. For each distance between $i$ and $j$, we learn a keypoint adaptive offset value. However, this $f(\mathbf{P})$ only works with the absolute position information of $\mathbf{P}$ and the relative distance between $i$ and $j$. It is missing the relative distance between $\mathbf{P}$ and ($i$, $j$).

\Paragraph{Relative $f(\mathbf{P})$}
To improve, $f(\mathbf{P})$ can be adjusted to work with the position of keys and queries \textit{relative} to the keypoints. In other words, 
so that the query-key relationship in $\mathbf{B}_{ij}$ depends on the query-landmark relationship. To achieve this, we generate a mesh grid $\mathbf{M}\in\mathbb{R}^{N\times 2}$ of patch locations containing all possible combinations of $i_x$ and $i_y$. $N$ represents the number of patches. We then compute the element-wise difference between the normalized grid and keypoints $\mathbf{P}$ to obtain a grid of $i,j$ relative to the keypoints:
\begin{equation}
\mathbf{D} = \text{Expand}(\mathbf{M}, \text{dim}\!=\!1) - \text{Expand}(\mathbf{P}, \text{dim}\!=\!0), 
\end{equation}
where $\mathbf{D}$ is the broadcasted tensor difference of shape $\mathbb{R}^{N\!\times\!N_L\!\times\!2}$ . Finally, we reshape $\mathbf{D}$ and linearly project it with $\mathbf{W}_L$. Specifically, 
\begin{align}
\mathbf{D}^{\prime} &=\text{Reshape}(\mathbf{D}) \in \mathbb{R}^{N\!\times\!(2N_L)}\\ 
f(\mathbf{P}) &= \mathbf{D}^{\prime}\mathbf{W}_L \in  \mathbb{R}^{N\times K}\\
\mathbf{B}_{ij} &= f(\mathbf{P})[i, d(i,j)] \in \mathbb{R}^1.
\end{align}

In other words, the offset value $\mathbf{B}_{ij}$ is determined with respect to the positions of the keypoints and is unique for each query location. This approach allows for more expressive control of the query-key relationships with the keypoint locations. An illustration of this is shown in Fig.~\ref{fig:figure3}.

\Paragraph{Multihead Relative $f(\mathbf{P})$}
Lastly, we can further enhance our method by tailoring the query-keypoint relationship for each head in the attention mechanism. When there are $H$ heads, we simply expand the dimension of $\mathbf{W}_L$ to $\mathbf{W}_L\in \mathbb{R}^{(2N_L)\times HK}$. By reshaping $f(\mathbf{P})$, we obtain $f(\mathbf{P})^h$ for each head. Furthermore, considering the multiple self-attentions in ViT which entails multiple RPEs, we can individualize $f(\mathbf{P})$ for each self-attention by additionally increasing the dimension of $\mathbf{W}_L$ to $\mathbf{W}_L\in \mathbb{R}^{(2N_L)\times N_dHK}$, where $N_d$ represents the transformer's depth. Since $f(\mathbf{P})$ is computed only once per forward pass, this modification introduces negligible computational overhead compared to other operations. In Sec.~\ref{sec:ablation}, we evaluate and compare the various KP-RPE versions (basic, relative keypoint, multiple relative keypoint), demonstrating the superior performance of the multiple relative keypoint approaches.

\section{Face Recognition Experiments}

\subsection{Datasets and Implementation Details}
To validate the efficacy of KP-RPE, we train our model using aligned face training data and evaluate on three distinct types of datasets: 1) aligned face data, 2) intentionally unaligned face data, and 3) low-quality face data containing misaligned images. For the evaluation, aligned face datasets include CFPFP~\cite{cfpfp}, AgeDB~\cite{agedb}, and IJB-C~\cite{ijbc}. 
For unaligned face data, we intentionally use the raw CFPFP~\cite{cfpfp} and IJB-C~\cite{ijbc} datasets without aligning them. Raw images, as provided by their respective creators, are equivalent to images cropped based on face detection bounding boxes. 
Lastly, we assess the model's robustness on low-quality face datasets, specifically TinyFace~\cite{tinyface} and IJB-S~\cite{ijbs}, which are prone to alignment failures.
This comprehensive setup enables us to examine the effectiveness of our proposed method across diverse data conditions.

\ablation
\ablationmodules

The training datasets MS1MV2~\cite{deng2019arcface} MS1MV3~\cite{deng2019lightweight} and WebFace4M~\cite{zhu2021webface260m} are released as aligned and resized to $112\!\times\!112$ by RetinaFace~\cite{deng2020retinaface} whose backbone is ResNet50 model trained on WiderFace~\cite{yang2016wider}. For keypoint detection in KP-RPE, we also use RetinaFace~\cite{deng2020retinaface} but with lighter backbone MobileNetV2 for faster inference. Given the sensitivity of ViTs to hyperparameters, we report the exact settings for learning rate, weight decay, and other parameters in the supplementary material. For ablation dataset, we take the MS1MV2 subset dataset as used in~\cite{kim2022adaface}.

Following the training conventions of~\cite{touvron2021training,kim2022adaface}, we adopt RandAug~\cite{cubuk2020randaugment}, repeated augmentation~\cite{hoffer2020augment}, random resized crop, and blurring. We utilize the AdaFace~\cite{kim2022adaface} loss function to train all models. For ablation, we employ ViT-small, while for SoTA comparisons, we use ViT-base models. The AdamW~\cite{adamw} optimizer and Cosine Learning Rate scheduler~\cite{loshchilov2016sgdr,rw2019timm} are used.
In WebFace4M trained models, we  adopt PartialFC~\cite{an2021partial,an2022killing} to reduce the classifier's dimension.

\subsection{Ablation Analysis}
\label{sec:ablation}

Row $1$ in Tab.~\ref{tab:ablation} shows results on the baseline ViT. Row $2$ and $3$ show results on the baseline ViT with iRPE and our proposed KP-RPE.  
KP-RPE demonstrates a substantial performance improvement on unaligned and low-quality datasets, without compromising performance on aligned datasets. Last row highlights the difference between ViT and ViT+KP-RPE. Also, Fig.~\ref{fig:interpolation} shows the sensitivity to the affine transformation, \textit{i.e.}, how the performance changes when one interpolates the affine transformation from the face detection images to the aligned images in CFPFP dataset. 

Tab.~\ref{tab:ablationtwo} further investigates the effect of modifications to KP-RPE. By making KP-RPE dependent on the difference between the query and keypoints (row $2$), we observe a significant improvement in unaligned dataset performance. Also, by allowing unique mapping for each head and module in ViT (row $3$), we achieve a further  improvement. In other words, more expressive KP-RPE is beneficial for learning complex RPE that depends on the keypoints of an image. Overall, the ablation study highlights the necessity of each component in KP-RPE and the effectiveness of KP-RPE in enhancing the robustness of face recognition models, particularly with unaligned and low-quality datasets.

\interoplation

\subsection{Computation Analysis}
\flop

In this section, we analyze the computational efficiency of our proposed KP-RPE in terms of FLOPs, throughput, and the number of parameters. 
Tab.~\ref{tab:flop} shows that KP-RPE is highly efficient, with only a small increase in the computational cost (FLOPs) compared to the backbone: $0.02$ GFLOP increase for ViT Small and 0.07 GFLOP increase for ViT Base (ViT vs ViT+KP-RPE). Notably, KP-RPE is considerably more efficient than iRPE, which incurs an increase of $0.71$ GFLOP for ViT Small and $1.42$ GFLOP for ViT Base.

Considering training throughput, which factors in computation time during training (with backpropagation), KP-RPE's efficiency is more pronounced. It only reduces throughput by $9.15$\% for ViT Small and $16.44$\% for ViT Base, as opposed to iRPE's larger decrease. Also, we show the GLOP and throughput with the landmark detection time included. Landmark detection time is negliable comparted to the total feature extraction time. 

Also, our method introduces a negligible increase in the number of parameters: just $0.05$M for ViT Small and $0.21$M for ViT Base. Hence, incorporating KP-RPE into the model achieves enhances performance without a substantial rise in computational cost or model complexity.

\subsection{Comparison with SoTA Methods}
\label{sec:sota}

In this section, we position ViT+KP-RPE, against SoTA face recognition methodologies with large-scale datasets and large models. We undertake a comprehensive evaluation, covering both high-quality and low-quality image datasets. The results, as shown in Tab.~\ref{SOTA_table}, underscore the strengths of KP-RPE. Notably, the inclusion of KP-RPE does not impair the performance on high-quality datasets, a testament to its applicability to both low and high-quality datasets.

This becomes particularly compelling when we observe the performance on low-quality datasets. Consistent with the findings of our ablation study, the introduction of KP-RPE leads to an appreciable improvement in these challenging scenarios. This supports our thesis that face recognition models with robust alignment capabilities can indeed enhance performance on low-quality datasets. In summary, our model with KP-RPE not only maintains competitive performance on high-quality datasets but also brings significant improvements on low-quality ones, marking it  a valuable contribution to the field of face recognition. 

\sotatable

\subsection{Note on the Landmark Predictor} 
KP-RPE in all experiments uses our own MobileNet~\cite{sandler2018mobilenetv2} based RetinaFace~\cite{deng2020retinaface} to predict landmarks for KP-RPE. We train MobileNet version for computation efficiency. However, the original landmark predictor used for aligning the test datasets is ResNet50-RetinaFace~\cite{deng2020retinaface}. We also report the KP-RPE performance with the officially released ResNet50-RetinaFace. We report this to compare KP-RPE on the same ground with other models by using the same landmark used to pre-align the testset. The face recognition performance of KP-RPE+Official is similar to KP-RPE+Ours ($75.86$ vs $75.80$ in TinyFace Rank1). Our MobileNet-RetinaFace is improved to perform similarly to ResNet50 in landmark prediction by applying additional tricks while training. Therefore, the face recognition performances are also similar. Unlike vanilla RetinaFace on face alignment, ours is fully differentiable during inference and name it Differentiable Face Aligner. Details and analysis can be found in Supp.2 and 3. 


\subsection{Scalability on Larger Training Datasets}

We train the ViT+KP-RPE model on a larger WebFace12M~\cite{zhu2021webface260m} dataset to demonstrate the potential of KP-RPE in its scalability and applicability in real-world, data-rich scenarios. Tab.~\ref{SOTA_table}'s last row shows that the performance continues to increase with WebFace12M dataset.

\Paragraph{Discussion} Why are noisy keypoints more useful in KP-RPE than in simple alignment?
The short answer is that not all predicted points are noisy in an image while alignment as a result of one or more noisy point impacts all pixels. For our attempt at a more detailed answer, please refer to Supp.2.4.

\section{Gait Recognition Experiments}

KP-RPE is a generic method that can generalize beyond face recognition to any task with keypoints. We apply KP-RPE to gait recognition using body joints as the keypoints.


\paragraph{Dataset.}
We train and evaluate on Gait3D~\cite{zheng2022gait}, an in-the-wild gait video dataset.
In our experiments, we use silhouettes and 2D keypoints preprocessed and released by the authors directly.
Following SMPLGait~\cite{zheng2022gait,ye2021deep}, we use rank-\(n\) accuracy (\(n=1,5,10\)), mean Average Precision (mAP), and mean Inverse Negative Penalty (mINP) for evaluation.

\Paragraph{Implementation Details}
We implement SwinGait-2D~\cite{fan2023exploring} as the baseline in our experiments. 
SwinGait-2D is chosen over SwinGait-3D~\cite{fan2023exploring} because we focus on exploiting the geometric information in gait recognition.
SwinTransformer~\cite{liu2021swin} uses vanilla relative positional encoding for each windowed self-attention. To incorporate KP-RPE into the SwinTransformer, we modify the 2D grid $\mathbf{M}$ to be the size of the window as opposed to the image size.
Following the default configuration of~\cite{zheng2022gait}, we use an AdamW~\cite{adamw} optimizer with a learning rate \num{3e-4} and weight decay \num{2e-2}, accompanied by an SGDR~\cite{loshchilov2016sgdr} scheduler.
We train our models for \num{60000} iterations, sampling \num{32} subjects and \num{4} sequences per subject in a batch.

\begin{table}
    \centering
    \footnotesize
    \caption{KP-RPE performance on Gait3D~\cite{zheng2022gait} compared with the baseline. KP-RPE boosts all metrics by a large margin.}
    \label{tab:gait_perf}
    \begin{tabular}{lcccc}
    \toprule
    Model & Rank-1 & Rank-5 & mAP & mINP \\ \midrule
    GaitSet~\cite{chao2019gaitset} & $36.7$ & $58.3$ & $30.01$ & $17.30$ \\
    MTSGait~\cite{mtsgait} & $48.7$ & $67.1$ & $37.63$ & $21.92$ \\
    DANet~\cite{danet} & $48.0$ & $69.7$ & --- & --- \\
    GaitGCI~\cite{dou2023gaitgci} & $50.3$ & $68.5$ & $39.5$ & $24.3$ \\
    GaitBase~\cite{fan2022opengait} & $64.6$ & $81.5$ & $55.31$ & $31.63$ \\ 
    HSTL~\cite{hstl} & $61.3$ & $76.3$ & $55.48$ & $34.77$ \\ 
    DyGait~\cite{DyGait} & $66.3$ & $80.8$ & $56.40$ & $\mathbf{37.30}$ \\ \midrule
    SwinGait-2D~\cite{fan2023exploring} & $67.1$ & $83.7$ & $58.76$ & $34.36$ \\
    \hspace{5pt} + KP-RPE & $\mathbf{68.2}$ & $\mathbf{84.4}$ & $\mathbf{60.81}$ & $36.19$ \\ 
    \bottomrule
    \end{tabular}
\end{table}

\Paragraph{Results and Analyses} In Tab.~\ref{tab:gait_perf}, we compare to SoTA approaches, including SwinGait-2D~\cite{fan2023exploring}, with and without KP-RPE.
We can see that the KP-RPE shows a significant improvement over SwinGait-2D, with \qty{1.1}{\percent} and \qty{0.7}{\percent} improvement on rank-$1$ and -$5$ accuracies, respectively.
mAP has improved by \qty{2.05}{\percent} and mINP by \qty{1.23}{\percent} of the baseline) compared to SwinGait-2D.
We believe that a great portion of the improvement comes from KP-RPE exploiting the gait information contained in 2D skeletons.
Gait skeletons contain identity-related information, such as body shape and walking posture.
This demonstrates that KP-RPE is both effective and generalizable to the gait recognition.


\section{Conclusion}
In this work, we introduce Keypoint-based Relative Position Encoding (KP-RPE), a method designed to enhance the robustness of recognition models to alignment errors. Our method uniquely establishes key-query relationships in self-attention based on their distance to the keypoints, improving its performance across a variety of datasets, including those with low-quality or misaligned images. KP-RPE demonstrates superior efficiency in terms of computational cost, throughput and recognition performance, especially when affine transform robustness is beneficial. We believe that KP-RPE opens a new avenue in recognition research, paving the way for the development of more robust models.

\Paragraph{Limitations}
While  KP-RPE  shows impressive face recognition capabilities, it does require keypoint supervision, which may not always be readily available and can constrain its application, particularly when the dataset does not comprises of images with a consistent topology. Future work should consider the self-discovery of keypoints to lessen this dependence, thereby boosting the model's flexibility.

\paragraph{Potential Societal Impacts}
Within the CV/ML community, we must strive to mitigate any negative societal impacts. This study uses the MS1MV* dataset, derived from the discontinued MS-Celeb, to allow a fair comparison with SoTA methods. However, we encourage a shift towards newer datasets, showcasing results using the recent WebFace4M dataset. Data collection ethics are paramount, often requiring IRB approval for human data collection. Most face recognition datasets likely lack IRB approval due to their collection methods. We support the community in gathering large, consent-based datasets or fully synthetic datasets~\cite{bae2023digiface1m,kim2023dcface}, enabling research without societal backlash.

\Paragraph{Acknowledgments} This research is based upon work supported by the Office of the Director of National Intelligence (ODNI), Intelligence Advanced Research Projects Activity (IARPA), via 2022-21102100004. The views and conclusions contained herein are those of
the authors and should not be interpreted as necessarily representing the official policies, either expressed or implied, of ODNI, IARPA, or the U.S. Government. The U.S. Government is authorized to reproduce and distribute reprints for governmental purposes notwithstanding any copyright annotation therein.

\clearpage
{
\small
\bibliographystyle{plain}
\bibliography{bibliography}
}

\clearpage


\onecolumn
\clearpage
\begin{center}
    \LARGE{\textbf{KeyPoint Relative Position Embedding for Face Recognition}}\\[10pt]
    \normalsize{Supplementary Material}
\end{center}
\setcounter{page}{1}
\setcounter{section}{0}
\section{Training Details}

Training code will be released for reproducibility. Our experiments were conducted using the PyTorch deep learning framework.  Detailed information pertaining to the training parameters, configurations, and specifics can be referred to in Tab.~\ref{tab:traindetail}. We employed the Vision Transformer (ViT) model architectures as implemented in the InsightFace GitHub repository, ensuring a well-established and tested model foundation. When measuring the throughput of our KeyPoint Relative Position Embedding (KPRPE), we utilized an NVIDIA RTX3090 GPU.

\begin{table}[h]
\caption{Details for training face recognition models with or without KPRPE.}
\vspace{2mm}
\small
\centering
\begin{tabular}{ccc}
                    & \textbf{Ablation Experiments}                                               & \textbf{Large Scale Experiments }                                                 \\\hline
Backbone            & ViT Small                                              & ViT Large                                                      \\\hline
LR                  & 0.001                                                  & $0.0001$                                                         \\\hline
Batch Size                  & $512$                                                  & $1024$                                                         \\\hline
Epoch               & $34$                                                     & $36$                                                             \\\hline
Momentum            & \multicolumn{2}{c}{$0.9$}                                                                                                 \\\hline
Weight Decay        & \multicolumn{2}{c}{$0.05$}                                                                                                \\\hline
Scheduler           & \multicolumn{2}{c}{Cosine}                                                                                              \\\hline
Optimizer           & \multicolumn{2}{c}{AdamW}                                                                                               \\\hline

Warmup              & \multicolumn{2}{c}{$3$}                                                                                                   \\\hline
AdaFace Loss Margin & \multicolumn{2}{c}{$0.4$}                                                                                                 \\\hline
AdaFace Loss $h$      & \multicolumn{2}{c}{$0.333$}                                                                                               \\\hline
Augmentation        & \multicolumn{2}{c}{\makecell{Flip, Brightness, Contrast, Scaling, \\Translation, RandAug~\cite{cubuk2020randaugment}(magnitude:14/31), \\Blur, Cutout, Rotation ($20^{\circ}$)}} \\\hline
PartialFC           & None                                                   & sampling rate $0.6$  \\\hline
RepeatedAug Prob                 & $0.5$                                                  & $0.1$                                                         \\\hline
\end{tabular}
\label{tab:traindetail}

\end{table}

\newpage
\section{Supplementary Performance Analysis}

\subsection{Performance Across Various Loss Functions}

In our extensive evaluation, we have employed three popular loss functions: AdaFace~\cite{kim2022adaface}, CosFace~\cite{wang2018cosface}, and ArcFace~\cite{deng2019arcface}, to train the Vision Transformer (ViT) in combination with our proposed KeyPoint Relative Position Embedding (KPRPE). As demonstrated by the results in Tab.~\ref{tab:additional} rows 3-6, our method exhibits consistent performance improvements on lower quality datasets across all three loss functions when compared to the standalone ViT. This signifies the versatility of KPRPE in synergizing with a variety of loss functions to enhance the robustness of face recognition models to less-than-optimal image quality.

\begin{table*}[h]
\centering

\caption{SoTA comparison on low-quality and high-quality datasets. IJB-C~\cite{ijbb} reports TAR@FAR=0.01\%}
\vspace{-1mm}
\footnotesize
\label{SOTA_table2}
\scalebox{1.0}{
\begin{tabular}{lccccccccc}
\hline
\multicolumn{1}{c}{\multirow{3}{*}{Method}} & \multicolumn{1}{c}{\multirow{3}{*}{Backbone}} & \multicolumn{1}{c}{\multirow{3}{*}{Train Data}}  &\multicolumn{4}{c}{Low Quality Dataset}   &\multicolumn{3}{c}{High Quality Dataset} \\ 
   &  &  & \multicolumn{2}{c}{TinyFace~\cite{tinyface}}   & \multicolumn{2}{c}{IJB-S~\cite{ijbs}}    &  AgeDB  &  CFPFP &  IJB-C  \\
           &   &   &   \multicolumn{1}{c}{Rank-$1$} & \multicolumn{1}{c}{Rank-$5$} & \multicolumn{1}{c}{Rank-$1$} & \multicolumn{1}{c}{Rank-$5$}  & \multicolumn{2}{c}{\scalebox{0.875}{Verification Accuracy}} & \scalebox{0.875}{0.01\%} \\ \hline
AdaFace~\cite{kim2022adaface}   &  \textbf{ViT}    &  \scalebox{0.86}{WebFace4M~\cite{zhu2021webface260m}}    & $74.81$ &  $77.58$ & $71.90$  & $77.09$  & $97.48$  &  $98.94$ & $97.14$ \\    
AdaFace~\cite{kim2022adaface}   &  \textbf{ViT+KPRPE}    &  \scalebox{0.86}{WebFace4M~\cite{zhu2021webface260m}}   &  $\mathbf{75.80}$  &  $78.49$   & $72.78$  & $78.20$  & $\mathbf{97.67}$ & $99.01$ &  $97.13$ \\
ArcFace~\cite{deng2019arcface}   &  \textbf{ViT+KPRPE}    &  \scalebox{0.86}{WebFace4M~\cite{zhu2021webface260m}}   &  $75.62$ &    $\mathbf{78.57}$  & $\mathbf{73.04}$   &  $\mathbf{78.62}$ & $97.57$  & $\mathbf{99.06}$  & $\mathbf{97.21}$\\  
CosFace~\cite{wang2018cosface}   &  \textbf{ViT+KPRPE}    &  \scalebox{0.86}{WebFace4M~\cite{zhu2021webface260m}}   & $75.48$ & $78.30$  & $72.22$  & $77.67$  & $97.45$  & $98.94$  & $96.98$ \\\hline
\end{tabular}
}
\label{tab:additional}

\end{table*}

\subsection{Performance with Different Number of Keypoints}

We include the impact of the number of keypoints in KP-RPE. We initiated the analysis with 5 keypoints, the maximum available in RetinaFace. And gradually reduce the number of points.

\begin{table}[h]
\centering
\footnotesize
\begin{tabular}{ccccc}
\hline
Number of Keypoints & TinyFace Rank1 & TinyFace 5 & AgeDB & CFPFP \\\hline
5 & $\bm{69.88}$ & $\bm{74.25}$ &$\bm{ 95.92}$ & $96.60$ \\\hline
4 & $69.58$ & $73.63$ & $95.65$ & $96.57$ \\\hline
3 & $69.66$ & $73.95$ & $95.77$ & $\bm{96.80}$ \\\hline
2 & $69.26$ & $73.42$ & $95.73$ & $95.97$ \\\hline
No Keypoints (Vanilla ViT) & $68.24$ & $72.96$ & $95.57$ & $96.11$ \\\hline

\end{tabular}

\end{table}

For datasets characterized by lower image quality like TinyFace, the performance diminishes as the number of keypoints reduce. But it does not diminish compared to not using the keypoints. It could be that the information about the scale and rotation of an image could still be captured by few points as 2 or 3. 
Interestingly, in high-resolution datasets, the trend is absent and the performance remains relatively consistent regardless of the number of keypoints used. More keypoints can be adopted with other landmark detectors but they are not trained with low quality images in WiderFace as the dataset only provides 5 points.

\subsection{Sensitivty to Landmark Error in KPRPE}
\begin{wrapfigure}[17]{r}{0.4\textwidth}
\vspace{-6mm}
\centering
  \includegraphics[width=0.4\textwidth]{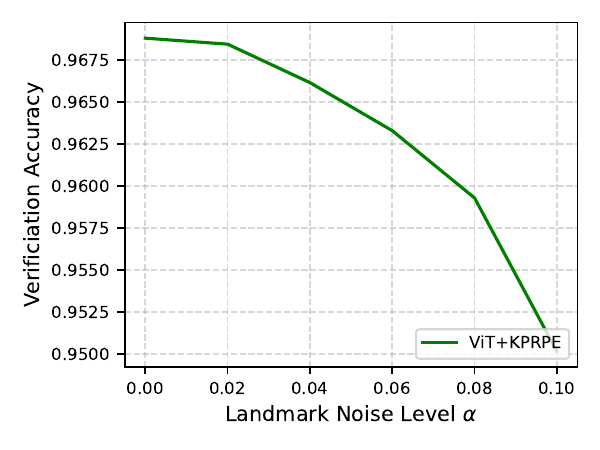}
  \caption{Verification accuracy measured in CFPFP dataset with added noise in landmark predictions. }
  \label{fig:ldmkinterpolate}
\end{wrapfigure}

To test the sensitivity of KPRPE to the landmark prediction error, we take the prediction of the landmark predictor and perturb it by the following equation, 

\begin{equation}
    \mathbf{L}_{pert} = \mathbf{L} + \alpha \mathbf{L}. 
\end{equation}

$\alpha$ is a parameter that changes the level of noise in the prediction. We change $\alpha$ from $0$ to $0.1$ after noting that $0.1$ makes the NME score to be $0.12$ which is far worse than the NME score of $0.05$ in WiderFace which is a harder dataset. Therefore, $\alpha=0.1$ is an extreme scenario where all of the inputs have failed to the level which exceeds the average level of failure in WiderFace by two times. 

Note that as we add noise into the landmark prediction, the performance goes down, signaling that KPRPE is dependent on the landmark prediction. However, the amount of performance drop within the range of realistic noise level is not too much (about 1.5\%). Fig~\ref{fig:pipeline} shows the experiment setting in a diagram.

\subsection{Why are noisy keypoints more useful in KP-RPE than in simple alignment?}
\begin{figure}[h]
    \centering
    \includegraphics[width=0.4\linewidth]{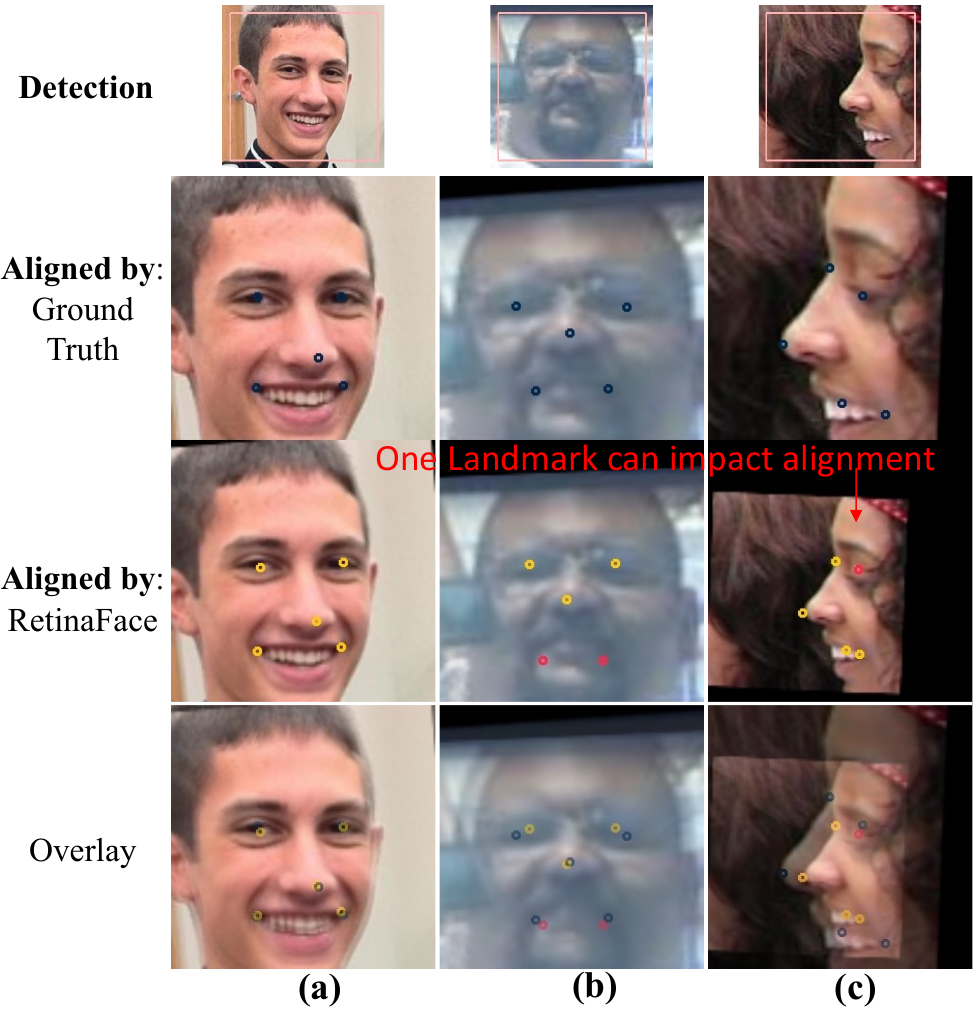}
    \vspace{-3mm}
    \caption{Keypoints in Aligned images. Blue: ground truth keypoints. Yellow/Red: RetinaFace keypoints with less/more than 5\% error from GT. Overlay of (b,c) shows how small deviation from one or two points can lead to significant scale, translation change. }
    \label{fig:misalign}
\end{figure}

The short answer is that not all predicted points are noisy in an image while alignment as a result of one or more noisy point impacts all pixels. For a more concrete example, in Fig.~\ref{fig:misalign}, we have taken images from WiderFace which contains human-annotated ground truth keypoints and compared them with RetinaFace prediction. Fig.~\ref{fig:misalign} (a) shows a well aligned scenario. (b) and (c) show that when one or two landmarks (red color) deviate from the ground truth (GT), the resulting alignment changes dramatically. For KP-RPE, this is a less severe problem because individual landmarks affect the RPE {\it indepenently} in the landmark space (0-1). On the other hand, when affine transformation is regressed to align the image to a canonical space, individual landmark error becomes {\it correlated} and {\it amplified}.     

\newpage
\section{Training Landmark Detector (MobileNet-RetinaFace)}
\label{sec:sfd}

RetinaFace~\cite{deng2020retinaface}, a single-stage face detector, is built upon Feature Pyramid Network~\cite{fpn} (FPN) and Single Shot MultiBox Detector~\cite{liu2016ssd}. It is originally designed for detecting multiple faces using anchor boxes in each location in an image. However, in our case, we assume the presence of one face, and we leverage this constraint to improve the landmark detection performance and efficiency of the model. This assumption is valid if a face detector crops out a face, which is a standard practice in face recognition. With this assumption, we can modify the RetinaFace to predict more accurate landmarks when the input image is cropped. We adopt few training techniques and a faster aggregation technique and name it Differentiable Face Aligner (DFA). The name suggests that with the modifications we propose, the face alignment network is differentiable (unlike RetinaFace because of NMS and CPU based cropping), making it potentially useful for other applications in computer vision. 

\paragraph{Training Data Adaptation}
We adapt the training data WiderFace~\cite{yang2016wider} for our Differentiable Face Aligner (DFA) by cropping out facial images using the ground truth bounding boxes. And we resize the input to be 160x160. This change in data size and distribution allows the model to specialize in localizing landmarks for single faces, ultimately improving its performance.

\paragraph{Aggregation Network}
The motivation for the aggregation network is to eliminate the  Non-Maximum Suppression (NMS) and output a single landmark prediction from multiple anchor boxes. We design a network that takes in the output of FPN and aggregates it to a single prediction.  The architecture of the aggregation network consists of MixerMLP~\cite{tolstikhin2021mlp}.
Specifically, let $\mathbf{X}$ be an image, and let $\mathbf{F}_{\textit{bbox}}$, $\mathbf{F}_{\textit{score}}$ and $\mathbf{F}_{ldmk}$ be the set of the output of FPN followed by the corresponding multitask head (bounding box, face score and landmark prediction) for each anchor box. For example, when an image is sized $160\times 160$, there are $1050$ anchor boxes. Based on these outputs, we predict the weights for fusing the outputs. Specifically, 
\begin{align}
    \mathbf{O} & = \text{Concat}(\mathbf{F}_{\textit{bbox}}, \mathbf{F}_{\textit{score}}, \mathbf{F}_{\textit{ldmk}}) \in \mathbb{R}^{1050\times (C_{bbox}+C_{score}+C_{ldmk})}  = \mathbb{R}^{1050\times (4+1+10)},\\
    \mathbf{w} & = \text{Softmax}(\text{MixerMLP}(\mathbf{O})) \in \mathbb{R}^{1050}, \\
    \mathbf{L} & = \mathbf{w}^T \mathbf{F}_{\textit{ldmk}}.
\end{align}

The final output $\mathbf{L}$ is the weighted average of the landmarks in all anchor boxes. The aggregation network is trained end to end with the rest of the detection model with the smooth L2 Loss~\cite{ren2015faster} between $\mathbf{L}$ and the ground truth landmark  $\mathbf{L}^{GT}$.

By incorporating these modifications, we show in Sec.~\ref{sec:ldmkexp} that our DFA achieves superior landmark detection performance compared to the  RetinaFace while using a more efficient backbone architecture. 

\paragraph{Training Details}
For the training of our Differentiable Face Aligner (DFA), we incorporated specific training settings to optimize the performance. We used an input image size of $160$ pixels, with a batch size of $320$. Training was conducted for $750$ epochs, ensuring that the model had adequate exposure to learn and generalize from the dataset. Training was performed using the WiderFace training dataset, with images cropped using the ground truth bounding boxes and a padding of $0.1$.

\subsection{Landmark Detection Performance} 
\label{sec:ldmkexp}

\begin{wrapfigure}[18]{r}{0.4\textwidth}
\vspace{-\intextsep}
\centering
  \includegraphics[width=0.4\textwidth]{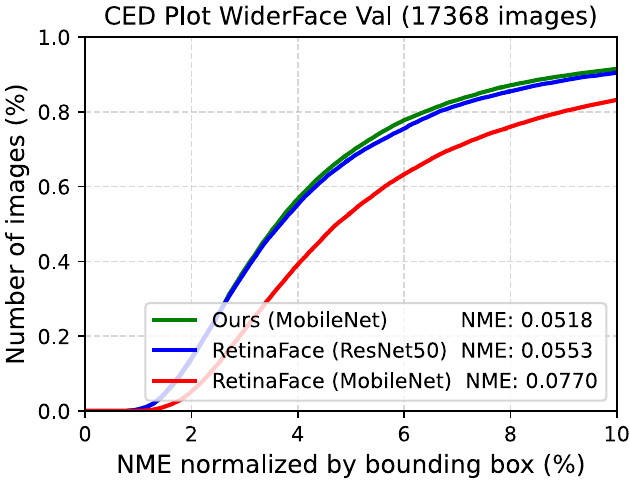}
  \caption{Cumulative Error Distribution curve and the corresponding NME for models evaluated on WiderFace~\cite{yang2016wider} validation set.  }
  \label{fig:cedplot}
\end{wrapfigure}

In this section, we evaluate the performance of our proposed Differentiable Face Aligner (DFA) in terms of landmark detection. We use the Normalized Mean Error (NME) as the metric and evaluate on WiderFace validation set~\cite{yang2016wider} as in RetinaFace~\cite{deng2020retinaface}. 

Fig.~\ref{fig:cedplot} shows an improvement in NME when using DFA compared to the baseline RetinaFace. The  RetinaFace  with MobileNet backbone achieves an NME of $0.077$, while the one with ResNet50  achieves $0.0553$. In contrast, our  DFA achieves $0.0518$, demonstrating its superiority in landmark detection.

Moreover, the DFA model benefits from the introduction of the aggregation network, which eliminates the need for the NMS stage. The improvement in NME due to the aggregation network is from $0.0527$ to $0.0518$. This not only simplifies the overall pipeline but also contributes to the enhanced performance of the DFA model in the landmark detection. With a straightforward modification in the training data and an aggregation stage that assumes a single-face image, a lightweight backbone with better performance can be trained. 

\section{IJB-S Evaluation Method}
IJB-S~\cite{ijbs} is a video-based dataset that defines probe and gallery templates according to its predefined video clip of arbitrary length. This naturally implies one must perform feature aggregation (fusion) when frame-level features are predicted. Since the backbone predicts a unit-norm feature vector for one image, the simplest method would be to average all the features within the template. The most popular method is to utilize norm-weighted average, where the features are averaged before normalization~\cite{kim2022adaface}. This only works if the norm is a good proxy for the prediction quality. However, in certain cases, depending on various factors such as dataset, learning rate, backbone, optimizer, etc., that go into the training of a model, this may not be the case. Also, in our experience, ViT+KPRPE was not the case. 

Therefore, we propose a proxy that could easily replace the norm with another quantity that can be found within a model. Since DFA predicts the landmarks $\mathbf{L}$ and a face score $\mathbf{F}_{score}$, we derive a fusion score using those quantities.  First, let us review the conventional norm-weighted feature fusion equation for a set of $N$ number of feature vectors $\{ f_i\}_N$ where $f_i = || f_{i} ||_2  \cdot  \bar{f}_i $ decomposes $f_i$ into the norm and the unit length feature. 
\begin{equation}
f_{\text{norm weighted}} = \frac{\sum_{i=1}^{N} || f_{i} ||_2 \cdot \bar{f}_{i}}{N}.
\end{equation}
In the equation above, ${f}_{i}$ represents the $i$-th frame-level feature, and $N$ is the total number of frames. Now, for KPRPE, we propose a new feature fusion method, incorporating the face score and the Euclidean distance between predicted landmarks $\mathbf{L}$ and the canonical landmark $\hat{\mathbf{L}}$, which is a known set of landmarks that the training images are aligned to. This distance score, $d$, is computed as:
\begin{equation}
d_i = \frac{h - \min(| \mathbf{L}_i - \hat{\mathbf{L}} |_2, h)}{h},
\end{equation}
where $h=0.2$ is a fixed constant that allows the score to be bounded between $0$ and $1$.
The face score $\mathbf{F}_{score}^i$ represents the quality of the image, and $d_i$ assigns more weight to well-aligned images. Proposed feature fusion equation, hence, becomes:
\begin{equation}
f_{\text{KPRPE}} = \frac{\sum_{i=1}^{N} (d_i \cdot \mathbf{F}^i_{score}) \cdot \bar{f}_{i}}{N}.
\end{equation}
This method allows for the aggregation of features even when the feature norm does not serve as a good proxy for the quality of an image. In computing IJB-S result for ViT+KPRPE, we use this fusion method.

However, for a fair comparison in IJB-S, it is important to apply this fusion method to previous methods. Therefore, we include the breakdown of with and without landmark score based fusion. For single image based datasets such as TinyFace, AgeDB or CFPFP, feature fusion is not needed.

\begin{table}[h]
\centering
\footnotesize

\begin{tabular}{ccccc}
\hline
Training Data: MS1MV3 & Feature Fusion Method & IJBS Rank1 & IJBS Rank5 & TinyFace Rank1 \\\hline
ViT Base+IRPE & Average & $62.49$ & $70.50$ & $69.05$ \\
ViT Base+IRPE & Landmark based & $63.81$ & $71.30$ & $69.05$ \\\hline
ViT Base+KPRPE & Average & $63.44$ & $72.04$ & $69.88$ \\
ViT Base+KPRPE & Landmark based & $64.68$ & $72.33$ & $69.88$ \\\hline

\end{tabular}

\begin{tabular}{ccccc}
\hline
Training Data: WebFace4M & Feature Fusion Method & IJBS Rank1 & IJBS Rank5 & TinyFace Rank1 \\\hline
ViT Large+IRPE & Average & $71.32$ & $76.22$ & $74.92$ \\
ViT Large+IRPE & Landmark based & $71.93$ & $77.14$ & $74.92$ \\\hline
ViT Large+KPRPE & Average & $65.95$ & $71.64$ & $75.80$ \\
ViT Large+KPRPE & Landmark based & $72.78$ & $78.20$ & $75.80$ \\\hline

\end{tabular}
\caption{Breakdown of with and without fusion method in various backbones and datasets. }
\end{table}

The performance of ViT+KP-RPE consistently surpasses ViT+iRPE, both in scenarios using Averaging or Landmark-based fusion. This affirms the efficacy of KP-RPE in enhancing performance, even in single image contexts like TinyFace. Importantly, while the keypoint detection step is integral to KP-RPE, it isn't incorporated within iRPE, making a direct comparison based on this score less straightforward for iRPE.

Interestingly, average fusion does not synergize well with ViT+KP-RPE. Contrary to typical observations where feature magnitude positively correlates with image quality~\cite{kim2022adaface}, with ViT+KP-RPE, a higher feature magnitude actually suggests reduced image quality. It remains unclear why this inverse relation emerges in our model. Through empirical observations, the relationship between feature magnitude and image quality appears contingent on the chosen training dataset and model architecture. For instance, models based on the ResNet architecture consistently exhibit a positive correlation between feature magnitude and image quality.

\section{Alignment Visualizations}
\begin{figure}[h]
    \centering
    \includegraphics[width=0.3\textwidth]{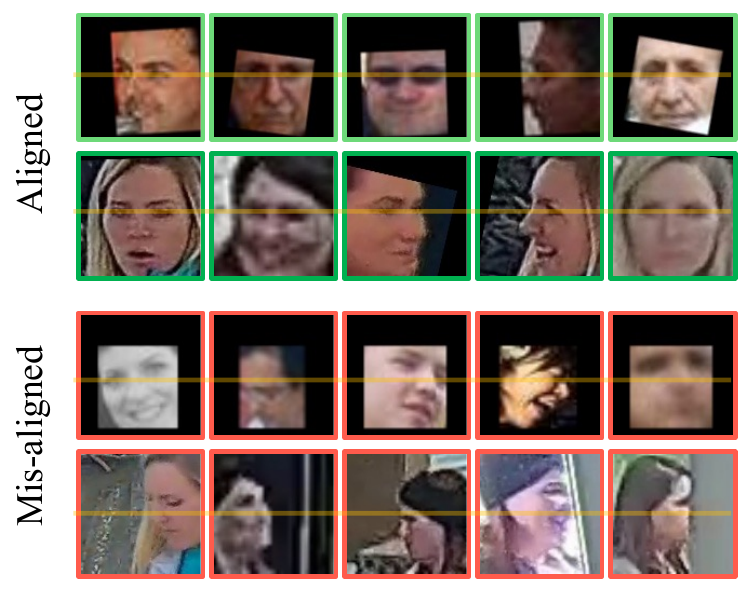}
  \caption{Actual examples of aligned and mis-aligned images from TinyFace~\cite{tinyface} (row1,3) and IJB-S~\cite{ijbs} (row2,4) datasets. These are shown as processed and used by~\cite{kim2022adaface}. Lines are placed on the eyes for a visual guide for an alignment.}
  \label{fig:lqexample}
\end{figure}

TinyFace~\cite{tinyface} and IJBS~\cite{ijbs}, which are prone to alignment failures. In Fig~\ref{fig:lqexample} we show some success and failure caes in alignment. These images are taken from the released aligned dataset itself. 

\section{Comparison with SoTA Off-the-Shelf Landmark Detector}
We evaluate the off-the-shelf landmark detector SLPT~\cite{xia2022sparse} (CVPR2022), which delivers strong performance on the high-quality WFLW~\cite{wu2018look} dataset. However, its performance dips significantly on the WiderFace dataset, populated with lower-quality images, as demonstrated in Tab.~\ref{tab:slpt}. This evaluation is not aimed at drawing a direct comparison between SLPT and DFA, as DFA is trained specifically on WiderFace. Instead, it serves to underline the performance variations of landmark detectors when trained on diverse datasets, stressing the importance of training dataset selection. Additionally, DFA boasts a magnitude faster speed than SLPT. 

\begin{figure}[h]
    \centering
    \includegraphics[width=0.5\linewidth]{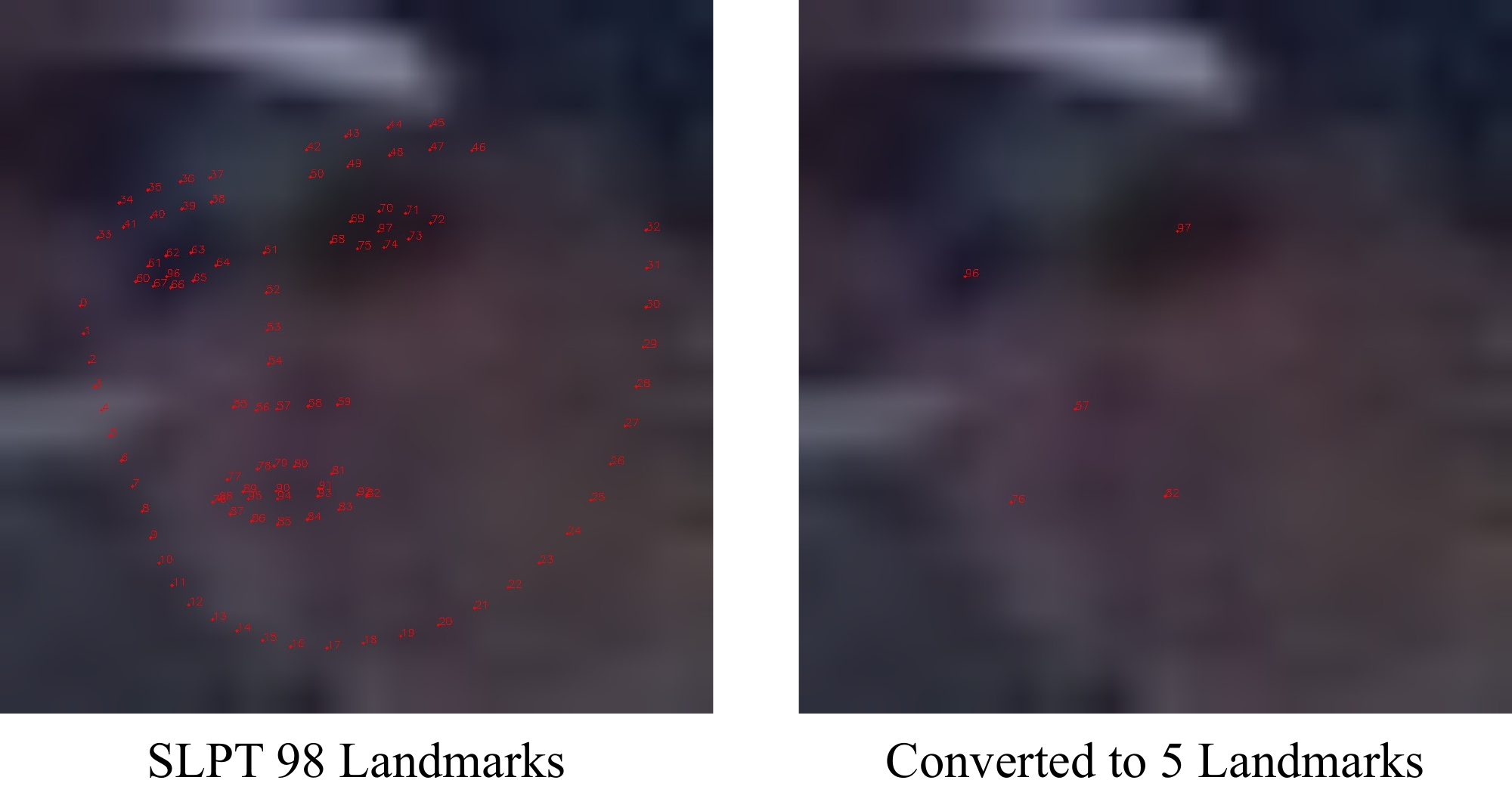}
    \caption{For converting 98 points landmarks from SLPT output, we choose indices 96, 97, 57, 76, 82. }
    \label{fig:slpt}
    
\end{figure}

\begin{figure}[h]
    \centering
    \includegraphics[width=\linewidth]{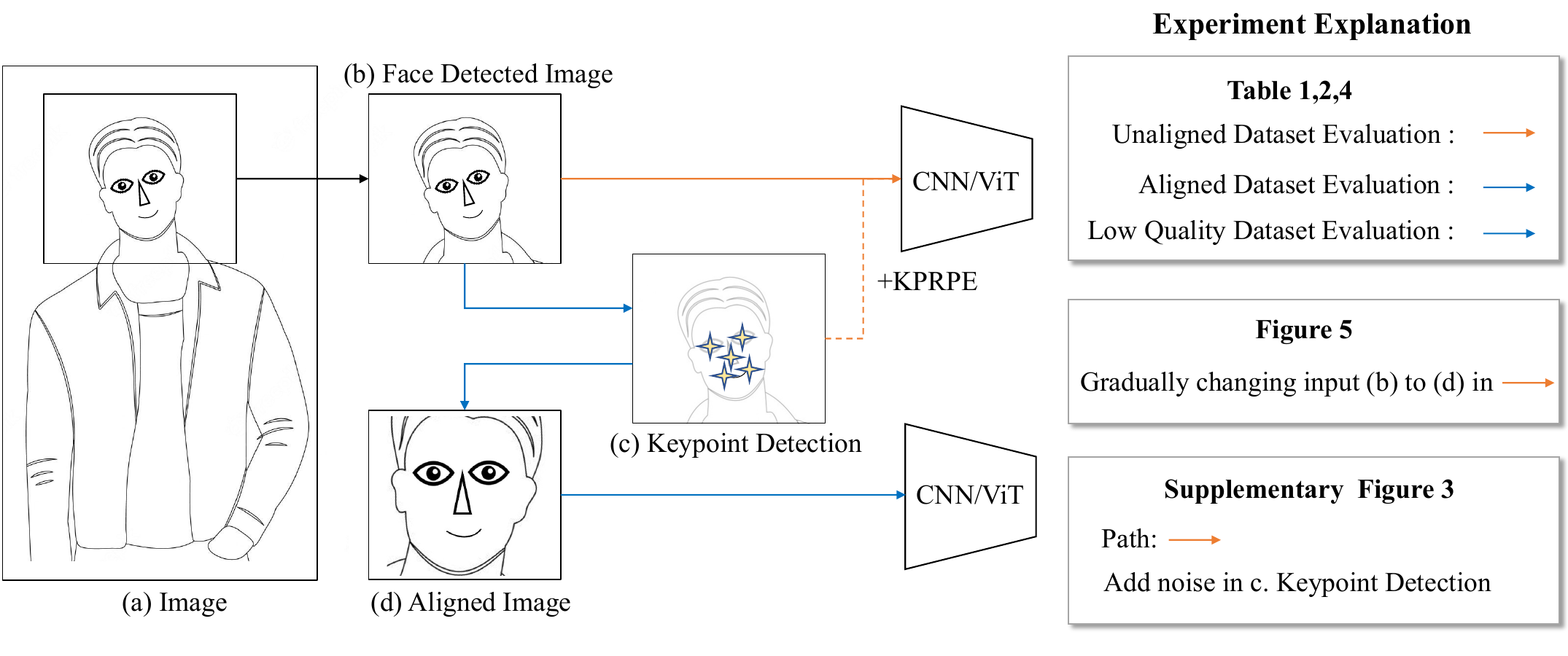}
    \caption{An illustration of face recognition pipeline from the raw image (a) to the aligned image (d).  }
    \label{fig:pipeline}
\end{figure}

Since SLPT predicts $98$ landmarks compared to $5$ landmarks in DFA, we convert the SLPT landmarks by selecting indices that represent the left eye, right eye, nose, left mouth, and right mouth. An example is shown in Fig.~\ref{fig:slpt}.

\begin{table}[!ht]
    \centering
    \caption{Comparison of DFA to SoTA Landmark detector. Note that NME is evaluated on on WiderFace Validation set. DFA is trained on WiderFace training set. SLPT is trained on WFLW. Direct NME comparison is not fair as the training dataset is different.}
    
    \begin{tabular}{ccccc}
    \hline
        Models & Train Data & NME  & FLOP & Params \\ \hline
        DFA MobileNet & WiderFace~\cite{yang2016wider} & $0.0518$ & $0.14$ GFLOP & $0.49$M \\
        SLPT~\cite{xia2022sparse} 6 Layer & WFLW~\cite{wu2018look} & $0.1104$ & $8.40$ GFLOP & $13.19$M \\ \hline
    \end{tabular}
    \label{tab:slpt}
\end{table}

\newpage
\section{Pipeline Detail}
In this section, we elaborate on the inference scenarios involved in evaluation pipelines. A face recognition pipeline could be simplified to the following diagram. For a given raw image (a), the face detector crops out an image containing a face region (b). Then a conventional alignment algorithm (MTCNN, RetinaFace, DFA) simultaneously predicts the landmarks (c) from (b). The least-square minimization algorithm is used to align (b) into the aligned image (d) using keypoints (c) and a reference landmark. This reference landmark is arbitrarily chosen, but the FR community usually adopts one popular setting. 

When one trains or evaluates face recognition models, most of the time, it is using aligned images (d), highlighted by the blue path. In our main paper, Tables 1,2, and 4, the aligned dataset and low-quality dataset are evaluated this way. The unaligned dataset in Tables 1 and 2 refers to the orange path. Whenever KPRPE is used, the keypoints are predicted using the inputs (b) or (d) depending on the path.

\newpage
\section{KPRPE Visualization}

\begin{figure}[h]
    \centering
    \includegraphics[width=0.8\linewidth]{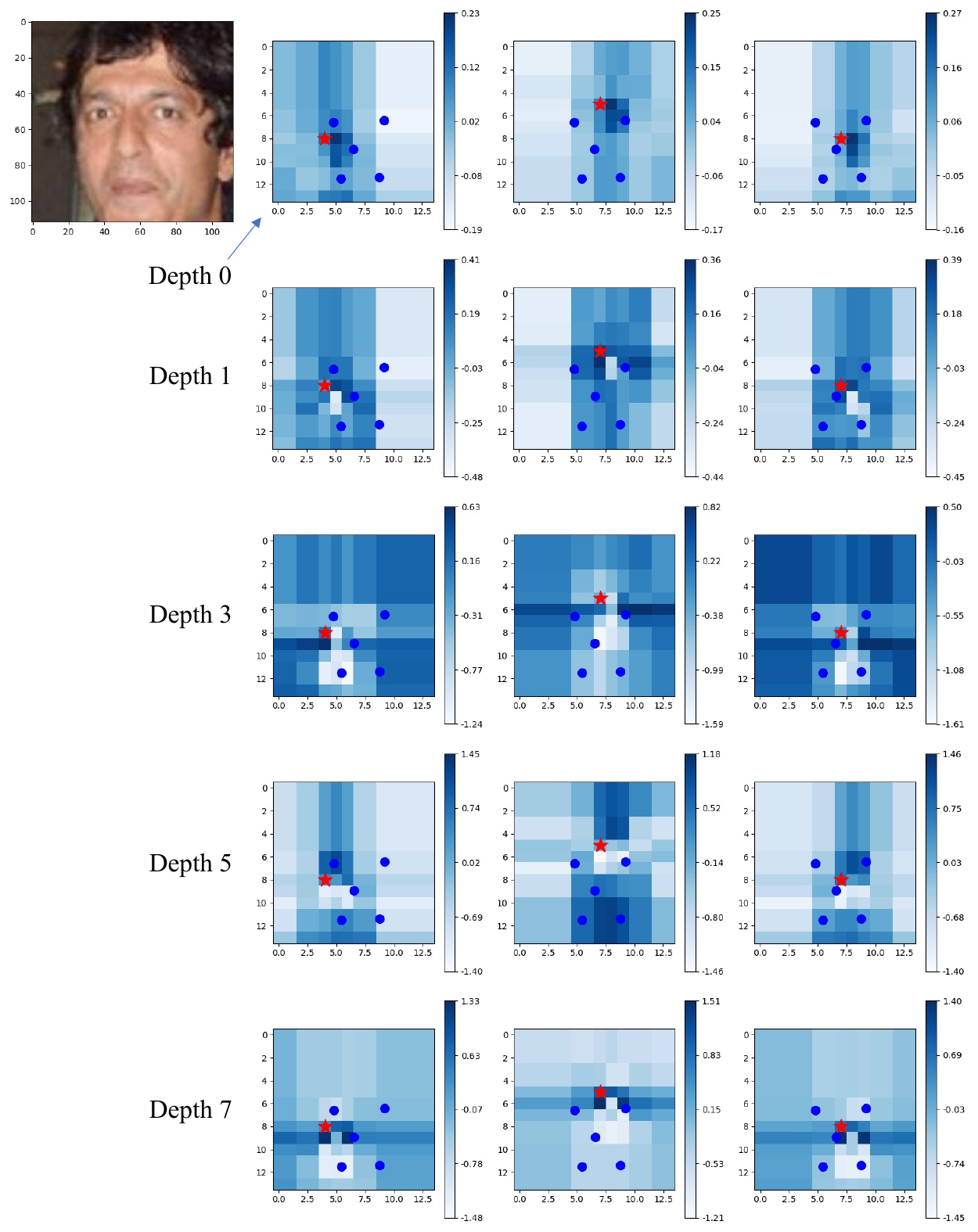}
    \caption{KPRPE Learned Offset $\mathbf{B}_{ij}$ visualization for different Transformer depths.}
    \label{fig:a1}
\end{figure}

We show the learned attention offset values in KPRPE. The red star denotes the query location and the blue circles represent the predicted landmarks. We pick head index 0 and plot the Transformer depth 0,1,3,5,7. Figs.~\ref{fig:a1} show different patterns of learned offset based on depth and query locations. Note that the higher values are denoted by a stronger blue color. Some attention offsets are 1) far from the query location, 2) horizontal pattern, etc. But there is an inherent bias toward attending nearby pixels. 

\begin{figure}[h]
    \centering
    \includegraphics[width=0.8\linewidth]{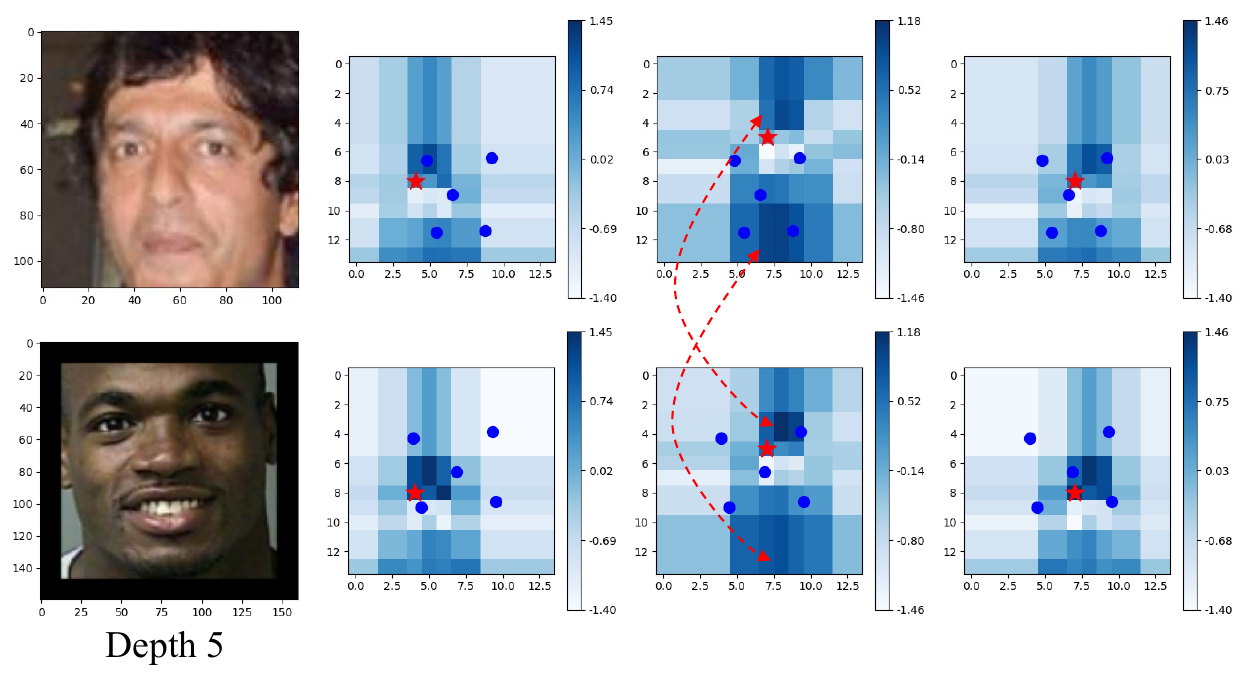}
    \caption{Cross image learned KPRPE visualization. We show the depth 5, and head index 0 of the same model. }
    \label{fig:a2}
\end{figure}

\newpage
Also, we show in Fig.~\ref{fig:a2}, an image with different images, therefore different landmark patterns. The changes in attention are not as dramatic as the changes across different head or depth. However, these changes observed in Fig~.\ref{fig:a2} account for the spatial variations in the image once they accumulate over all of the attention modules in the model.  

\end{document}